\documentclass[runningheads]{llncs}
\usepackage{graphicx}
\usepackage{tikz}
\usepackage{booktabs}       % professional-quality tables
\usepackage{comment}
\usepackage{amsmath,amssymb} % define this before the line numbering.
\usepackage{color}
\usepackage{multirow} 
\usepackage{amsfonts}       % blackboard math symbols
\usepackage{nicefrac}       % compact symbols for 1/2, etc.
\usepackage{microtype} % microtypography
\usepackage{cite}
\usepackage{caption} 
\usepackage{float}
\usepackage{varwidth}
\usepackage[utf8]{inputenc} 
\usepackage[T1]{fontenc} 
\usepackage[subrefformat=parens,labelformat=parens]{subcaption} %% provides subfigure
\usepackage{cleveref}

\usepackage[accsupp]{axessibility}  % Improves PDF readability for those with disabilities.

\begin{document}
% \linenumbers
\pagestyle{headings}
\mainmatter
\def\ECCVSubNumber{1011}  % Insert your submission number here

\title{Weakly Supervised Grounding for VQA in Vision-Language Transformers} % Replace with your title

%\begin{comment}
% \titlerunning{ECCV-22 submission ID \ECCVSubNumber} 
% \authorrunning{ECCV-22 submission ID \ECCVSubNumber} 
% \author{Anonymous ECCV submission}
% \institute{Paper ID \ECCVSubNumber}
%\end{comment}
%******************

% \begin{comment}
\titlerunning{Weakly Supervised Grounding for VQA in Vision-Language Transformers}
\author{Aisha Urooj Khan\inst{1}\orcidID{0000-0001-6521-2512} \and
Hilde Kuehne\inst{2,3}\orcidID{0000-0003-1079-4441} \and
Chuang Gan\inst{3}\orcidID{0000-0003-4031-5886} \and
Niels Da Vitoria Lobo\inst{1}\orcidID{0000-0001-5354-2805} \and
Mubarak Shah\inst{1}\orcidID{0000-0001-6172-5572}
}

\authorrunning{A. Urooj et al.} 

\institute{University of Central Florida, Orlando, FL, USA\\
\and
Goethe University Frankfurt, Frankfurt, Hesse, Germany \and
MIT-IBM Watson AI Lab, Cambridge, MA, USA
}

% \end{comment}
%******************6
\maketitle

\begin{abstract}
Transformers for visual-language representation learning have been getting a lot of interest and shown tremendous performance on visual question answering (VQA) and grounding. 
But most systems that show good performance of those tasks still rely on pre-trained object detectors during training, which limits their applicability to the object classes available for those detectors. 
To mitigate this limitation, the following paper focuses on the problem of weakly supervised grounding in context of visual question answering in transformers. 
% We combine the ideas of attention and capsules in a capsule-transformer architecture to improves grounding abilities of a vision-language transformers without the need of object masks as input for training. 
The approach leverages capsules by grouping each visual token in the visual encoder and uses activations from language self-attention layers as a text-guided selection module to mask those capsules before they are forwarded to the next layer. % \hkc{keep one of the two ... }
% For evaluation, we focus on the challenging GQA dataset for VQA grounding.
We evaluate our approach on the challenging GQA as well as VQA-HAT dataset for VQA grounding.
Our experiments show that: while removing the information of masked objects from standard transformer architectures leads to a significant drop in performance, the integration of capsules significantly improves the grounding ability of such systems and provides new state-of-the-art results compared to other approaches in the field\footnote{Code will be available at https://github.com/aurooj/WSG-VQA-VLTransformers}. 

% Transformers for visual-language representation learning have been getting a lot of interest and shown tremendous performance on various tasks including visual question answering (VQA) and grounding. 
% But most systems still rely on the object masks obtained from a pre-trained object detector during training. To address the task of weakly supervised grounding in context of visual question answer in transformers, the following paper proposes to leverage the concept of capsules in combination with self-attention layers by introducing a text-guided selection module.
% We propose to group each visual token via capsules in the feature dimension and use activations from language to mask those capsules before they are forwarded to the next layer. 
% This combines the ideas of attention and capsules in a capsule-transformer architecture that improves grounding abilities of a vision-language transformers without the need of object masks as input or for training. 
% % For evaluation, we focus on the challenging GQA dataset for VQA grounding.
% For evaluation, we focus on the challenging GQA dataset as well as VQA-HAT for VQA grounding.
% Our experiments show that, first, removing the information of masked objects leads to a significant drop of performance of the reference transformer architectures and second, how the integration of capsules helps to improve the grounding ability of such systems. 
\end{abstract}
% \input{teaser}
%---------------------------------------------------------------------------------------------
\begin{figure}[t]
\begin{center}

% \fbox{\rule{0pt}{2in} \rule{.9\linewidth}{0pt}}
 \includegraphics[width=\linewidth]{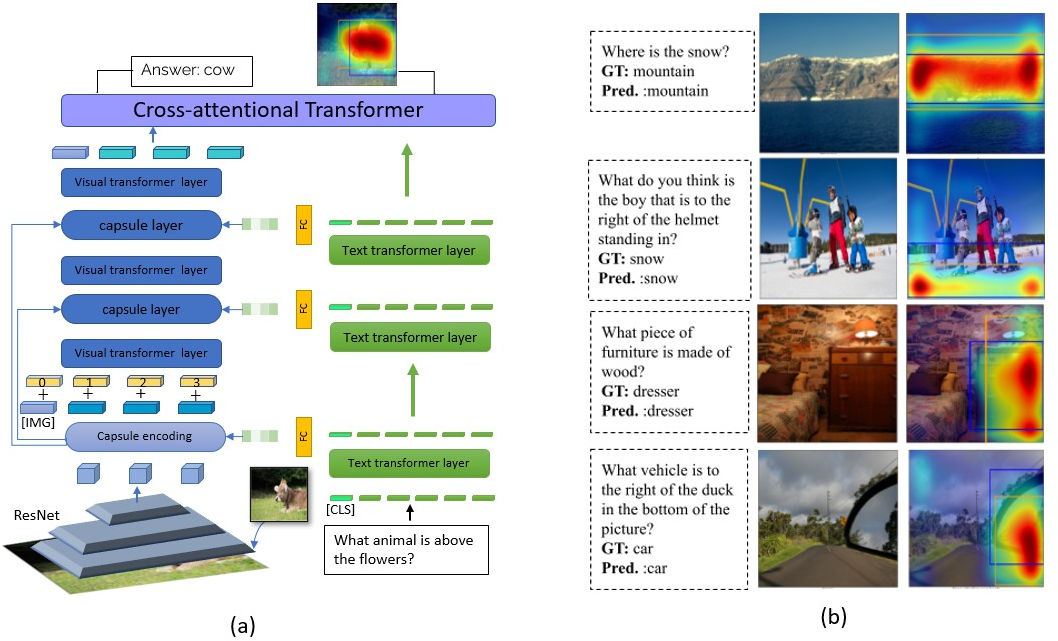}

\end{center}
\vspace{-5pt}
  \caption{ \small \textbf{(a) Proposed architecture}: Given the question-image pair, grid features are used to obtain visual capsules using a Capsule encoding layer. Output embedding for [CLS] token from text transformer layer is then used to do capsule features selection. Selected capsules encodings with position information is then input to the visual encoder. We use sentence embedding from each textual transformer layer to select capsules for each visual transformer layer. The selected capsules are then input to the next visual transformer layer along with the output of the previous layer in the visual encoder. Finally, a cross-attentional block allows for a fine-grained interaction between both modalities to predict the answer. \textbf{ (b) Attention from the proposed vision-language transformer with VQA supervision.} We look at the self-attention of the [IMG] token on the heads of the last layer. These maps show that the model automatically learns to ground relevant objects leading to weakly-supervised grounded VQA (GVQA). \textit{blue box is ground truth, orange is predicted box.}
  }
\vspace{-10pt}
\label{fig:data-samples}
\end{figure}

%--------------------------------------------------------------------------------

\section{Introduction}

\label{sec:intro}

Enabling VQA systems to be explainable can be important for variety of applications such as assisting visually-impaired people to navigate~\cite{gurari2018vizwiz, zeng2020vision} or helping radiologists  in early diagnosis of fatal diseases~\cite{abacha2019vqa, zhan2020medical}. A system that only produces a good answering accuracy will not be sufficient in these applications. Instead, VQA systems for such uses should ideally also provide an answer verification mechanism and grounding is a convincing way to obtain this direct verification.
% as e.g., grounding as that allows to directly verify the given answer.

Following upon success in natural language processing and multi-modal understanding, % transformers architectures have spread to many areas of computer vision tasks
a variety of transformer-based methods have been introduced for joint representation learning of vision and language and respective downstream tasks, including VQA. 
Such approaches, e.g.,~\cite{tan2019lxmert, vilbert12-in-1, li2019visualbert} are usually trained based on region masks of detected objects generated by a pre-trained object detector \cite{Khan2021TransformersSurvey}. 
The assumption that object masks are provided at input time limits detection to pretrained objects only and comes at the risk of image context information being lost. 
% that object detectors are usually trained for a finite set of classes, making the detection unreliable for unseen object classes 

%Detector-free transformers are simpler and faster because removing the extra step of extracting region-based features from a pre-trained object detector avoids expensive preprocessing as well as a bias toward pre-trained object classes. 
Detector-free methods avoid this bias toward pre-trained object classes while being simpler and faster because they do not need to extract region-based features from a pre-trained object detector. 
Other works~\cite{vilt, pixel-bert,desai2021virtex, clip} have therefore focused on removing the dependency on object detectors while achieving comparable if not better performance (e.g., on retrieval and VQA tasks). 
%Vision Transformer (ViT) \cite{dosovitskiy2020vit} was one of the first works showing successful application of transformers 
Among those, \cite{desai2021virtex} and \cite{clip} also show good visual representations by qualitative examples, but do not provide an evaluation of their answer (or question) grounding ability. 

% Evaluating transformers for their VQA grounding ability is essentially a weakly-supervised grounding setup because these systems use VQA supervision only. 

% Supervised VQA grounding requires bounding boxes 
% % (as input or training labels) 
% which is not scalable to real-world conditions. Weakly supervised VQA grounding, on the contrary, is more suited to operate under real-life conditions.
% % , but also challenging because we need to relate co-occurring visual entities with their textual counterparts.

In this work, we want to address this issue and focus on the problem of weakly supervised grounding for the VQA task in visual-language transformer based systems. Compared to detector-free referential expression grounding  \cite{xiao2017weakly1, chen2019weakly3, liu2021relationweakly4}, VQA breaks the assumption that the region description is always part of the input phrase as the answer word may not be present in the input question. It is therefore not enough to only learn a direct mapping between text and image features, but also requires processing multiple image-text mapping steps with the correct relation between them. \\
To address the task of VQA grounding in transformer-based architectures with the question-answering supervision alone, we propose a generic extension of the visual encoder part of a visual-language transformer based on visual capsule layers together with a text-guided selection module.  
Capsule networks learn to group neurons into visual entities. They thus try to capture entities (objectness) and relationships among them with promising results on various visual detection and segmentation tasks~\cite{duarte2018videocapsulenet, lalonde2018capsules, duarte2019capsulevos}.
%We combine the learned capsule representation with a selection module based on textual representations (question) from the language encoder to select features at entity-level, similar to attending object features, instead of and independent feature selection. \hkc{<-this sentence might be too much/duplicate?}
% Using capsules for weakly-supervised VQA grounding is therefore an intuitive choice.
% Capsules have already proven their value in aggregating visual features for detection and grounding in various setups, as e.g. in~\cite{duarte2018videocapsulenet}.
%
To make use of this ability in context of transformers, we transfer inputs as well as intermediate layers' feature tokens to capsule encodings, from which the most relevant ones will be selected by the textual tokens of a parallel language encoder. We propose to interleave transformer layers with such masked residual capsule encodings. This extension provides a combination of visual input routing and text-based masking which significantly improves the visual grounding ability of such systems. 

We evaluate existing methods as well as the proposed approach on the challenging GQA and VQA-HAT datasets. To this end, we 
%follow the protocol used by DINO \cite{dino} for object segmentation and 
consider the attention output obtained from these methods and evaluate it on various metrics, namely overlap, intersection over union, and pointing game accuracy.
Our results on the original architectures show a significant gap between task accuracy and grounding of existing methods indicating that existing vision-language systems are far from learning an actually grounded representation. The proposed method bridges the gap and outperforms the best contenders in terms of overlap, intersection over union, and pointing game accuracy achieving SOTA performance on the GQA dataset.  It also achieves best performance (in terms of mean-rank correlation score) on VQA-HAT\cite{vqahat} dataset among methods which do not use attention supervision. 

We summarize contributions of our architecture as follows: a) we propose a capsule encoding layer to generate capsule-based visual tokens for transformers; b) we propose a text-guided capsule selection with residual connections to guide the visual features at each encoding step; and c) we propose generic interleaved processing that can be integrated in various vision language architectures.

\section{Related Work}
\label{related_work}
\noindent \textbf{Visual-language Representation Learning.}
Learning a robust visual-language representation is currently an active area of research \cite{khan2021transformers} with impressive progress on downstream tasks including VQA \cite{lu2019vilbert, vilbert12-in-1, li2019visualbert, tan2019lxmert, chen2019uniter, miech2021thinking, li2020oscar,li2020unicoder, su2019vl}. A majority of these methods rely on object detections making the downstream task simpler. Some works have attempted to avoid this dependency on object detections and show comparable performance using spatial features or image patches \cite{ALBEF, vilt, soho, pixel-bert}. Our work also falls into the later category and uses grid features as input.

\noindent \textbf{Weakly-supervised Grounding and VQA.}
Weakly-supervised visual grounding is well studied for phrase-grounding in images \cite{xiao2017weakly1, chen2019weakly3, liu2021relationweakly4, datta2019align2groundweakly6, arbelle2021detectorweakly7, chen2018knowledgeweakly8, wang2021improvingweakly9}. Few works also focused on phrase grounding in videos \cite{shi2019notweakly2, yang2020weakly5, huang2018findingweakly10}. However, less attention has been paid to the VQA grounding despite having significance for many critical applications. There are many works on making questions visually grounded \cite{ramakrishnan2018overcominglg3, niu2021counterfactuallg4, selvaraju2019takinglg5, zhang2016automaticlg7, Whitehead_2021_CVPR, lang_groundlg8}, but only a handful of works are focused to evaluate their grounding abilities \cite{selvaraju2019takinglg5, hudson2019gqa, khan2021reason, riquelme2020explaininglg6, vqahat, qiao2018exploring}. 
% where \cite{riquelme2020explaininglg6, selvaraju2019takinglg5} focus on evaluating explanations using rank correlation between GradCAM attention and human attention maps. \cite{das2017humanlg1} show that using human attention maps make the VQA systems improve significantly but \cite{shrestha2020negativelg2} argued that the gains in performance was because of better regularization instead of using human attention. In that aspect, grounding evaluation on metrics such as overlap and IOU are a better indicator to assess systems for explanations.
GQA leverages scene graphs from Visual Genome dataset providing visual grounding labels for question and answer making it feasible to evaluate VQA logic grounding.
% To our knowledge, GQA is the only benchmark with image scene graphs which can be used to evaluate VQA logic grounding.
% GQA provides real-world compositional questions which require to look at different objects to predict the answer. 
Recently, xGQA \cite{pfeiffer2021xgqa} has been introduced as a multilingual version of GQA benchmark.
% providing questions and answers in 7 different languages. 
GQA \cite{hudson2019gqa} and \cite{khan2021reason} discuss the evaluation of VQA systems for grounding ability. VQA-HAT \cite{vqahat} on the other hand, provides human attention maps used for answering the question in a game-inspired attention-annotation setup. A handful of methods \cite{qiao2018exploring, lu2016hierarchical, yang2016stacked} evaluate their systems for correlation between machine-generated attention and human attention maps on VQA-HAT.
% and provides objects ground truths referenced in the question-answer pair. 
% However, GQA was not evaluated for grounding of VQA systems until very recently, when
% Recently, MAC-Capsules proposed by \cite{khan2021reason} focused on grounding of visual reasoning benchmarks. 
%We find this direction pointed out by MAC-Capsules interesting and important. 
However, with the emergence of transformers as the current SOTA, the focus moves towards grounding abilities of those systems for VQA task. Unfortunately, none of these transformer-based methods have yet focused on the evaluation of weakly supervised grounding. However, the fact that only few real-world dataset provide grounding labels makes this task challenging. We therefore finetune three existing detector-free transformer methods on GQA and evaluate them for the weak grounding task. 

\noindent \textbf{Transformers with Capsules.}
There are few works who focused on the idea of combining transformers and capsules 
\cite{mobiny2021transcaps1,gu2019improvingcaps2, pucci2021selfcaps3, wu2021centroidcaps4, liu2019transformercaps5, mazzia2021efficientcaps6, duan2020capsulecaps7}. For instance, \cite{wu2021centroidcaps4} studies text-summarization, image-processing, and 3D vision tasks; \cite{pucci2021selfcaps3, mobiny2021transcaps1} uses capsule-transformer architecture for image-classification, and \cite{liu2019transformercaps5} uses capsules-transformers for stock movements prediction. To the best of our knowledge, no one has studied the combination of capsules with transformers for VQA grounding. 
% Most of which are transformer-based methods. We therefore evaluate existing SOTA on transformer for answer grounding and propose a text-guided capsule-transformer method which outperforms the baseline systems including MAC-Capsules. 
 
\section{Proposed Approach}
\label{approach}
Given an input image-question pair with image $I$ and question $Q$, we want to localize the relevant question and answer objects with only VQA supervision. We start from a two stream visual-language model where the language encoder $L_e$ guides input and intermediate respresentations of the visual encoder $V_e$. The input text to the language encoder $L_e$ is a sequence of word tokens from a vocabulary $V$ appended with special tokens $[CLS]$ and $[SEP]$ at the start and end of the word tokens. As input to the visual encoder, our model takes convolutional features as image embeddings. The convolutional features $X \in \mathbb{R}^{h \times w \times d1}$ are extracted from a pre-trained ResNet model, $h, w$ are the feature height and width, and $d1$ is the extracted features dimension. A 2D convolutional layer then yields an embedding $X'$ of size $ \mathbb{R}^{h \times w \times d}$, where $d$ is the model dimension size. These input embeddings are used to produce capsule encodings $X_c$ as explained in section \ref{capsule_enc}.
%\hkc{We pretrain a capsule-based backbone on image-sentence pairs from MSCOCO \cite{} and Visual Genome \cite{} datasets. The pre-trained capsule-transformer is then fine-tuned for visual question answering task.} 
%Our approach extends this basic architecture by integrating capsules. 

In the following, we first explain the motivation to use capsules in Sec.~\ref{sec:capsulenetwork} followed by details about the capsule encoding in Sec.~\ref{capsule_enc}, the text-guided selection of the capsules in Sec.~\ref{masked_capsules}, as well as the text-based residual connection in Sec.~\ref{masked_residual}. We close the section with an overview of the pretraining procedure in Sec.~\ref{pretraining} and describe the details of the VQA downstream task in Sec.~\ref{VQADownstreamTask1}. 

% \begin{figure}[t]

% \begin{tabular}{c}
% {
% \includegraphics[width=0.5\linewidth]{images/capsule_enc_v3.png}
% }
% % \subcaptionbox{Self Attention}{\includegraphics[width=0.23\linewidth]{images/self-attention.pdf}} &
% % \subcaptionbox{Cross-Self Attention}{\includegraphics[width=0.4\linewidth]{images/cross-self-attention.pdf}} 
% \end{tabular}
% % \vspace{-10pt}
%  \caption{Overview of Capsule encoding layer: grid features $X' \in \mathbb{R}^{h\times w \times d}$ are transformed into capsules $X_c$ for each spatial position.
% %  of size  $\mathbb{R}^{h\times w \times d_c}$($d_c$=capsules dimension)
%  Output embedding $h^0_{cls}$ for $[CLS]$ token from the first text encoder layer generates a mask $m^0$ for capsule selection. The selected capsules $X^0_{mc}$ are flattened along capsule dimension to get a set of visual tokens (of length $h*w$) where each token is denoted by $x^0_{j_{mc}}, j=\{0,1,...,hw\}$; $x^0_{j_{mc}} \in \mathbb{R}^{d_c}$ where $ d_c$=capsule dimension, is then upsampled to model dimension $d$ using a fully connected layer. A position embedding is added to visual tokens with the special token [IMG] at position 0. The output capsule encodings are then input to the visual transformer for future steps.  
% %  Capsules obtained within capsule encoding layer also goes as input to the capsule layer (see figure \ref{fig:capsules-layer} for details).
%  }
% \label{fig:capsule-enc}
% \vspace{-10pt}
% \end{figure}

%----------------------------------------------------------------------------------------

\begin{figure}[t]

\begin{tabular}{cc}
% \multirow{2}{*}{\subcaptionbox{Proposed Architecture}{\includegraphics[width=0.45\linewidth]{images/overview-v5.png}}} &
\subcaptionbox{Capsule Encoding layer}{\includegraphics[width=0.5\linewidth]{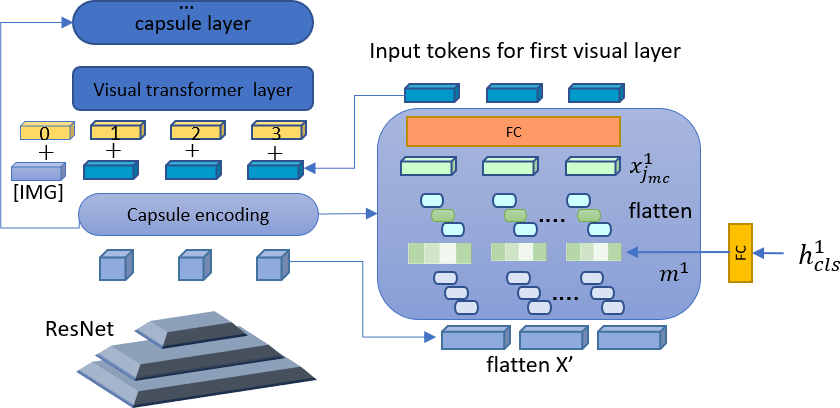}} &
\subcaptionbox{Capsule layer}{\includegraphics[width=0.5\linewidth]{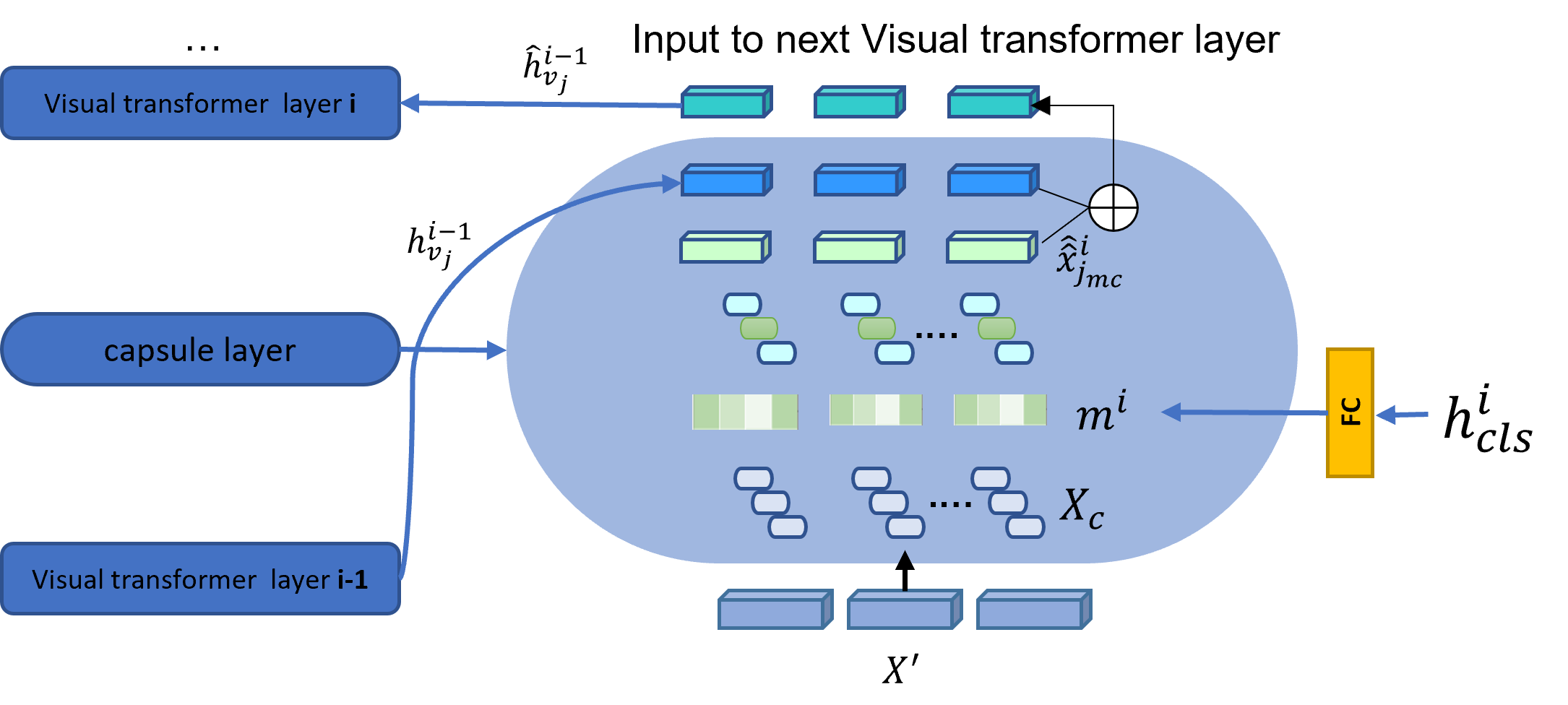}} 
\end{tabular}
\vspace{-5pt}
 \caption{\small  \textbf{(a) Capsule encoding layer}: grid features $X' \in \mathbb{R}^{h\times w \times d}$ are transformed into capsules $X_c$ for each spatial position.
%  of size  $\mathbb{R}^{h\times w \times d_c}$($d_c$=capsules dimension)
 Output embedding $h^1_{cls}$ for $[CLS]$ token from the first text encoder layer generates a mask $m^1$ for capsule selection. The selected capsules $X^1_{mc}$ are flattened along capsule dimension to get a set of visual tokens (of length $h*w$) where each token is denoted by $x^1_{j_{mc}}, j=\{1,2,...,hw\}$; $x^1_{j_{mc}} \in \mathbb{R}^{d_c}$ where $ d_c$=capsule dimension, is then upsampled to model dimension $d$ using a fully connected layer. A position embedding is added to visual tokens with the special token [IMG] at position 0. The output capsule encodings are then input to the visual transformer for future steps.  \textbf{(b) Capsule layer}: Similar to the Capsule encoding, input tokens $X'$ are first transformed into capsules $X_c$. The output feature $h^i_{cls}$ corresponding to $[CLS]$ token from the text encoder layer $i$ learns the presence probability of a certain capsule at the layer $i$ in visual encoder. This mask $m^i$ selects the capsules respective to the attended words at text encoder layer $i$. The resulting capsules are then flattened and upsampled (denoted by $\hat{x}^i_{j_{mc}}$) and added to the output $h^{i-1}_{cls}$ of the previous visual transformer layer $i-1$ to obtain input features $\hat{h}^{i-1}_{v_j}$ for the next visual transformer layer $i$.}
\label{fig:caps-arch}
\vspace{-10pt}
\end{figure}

\label{architecture}
% The full architecture is composed of modality-specific encoders with $L$ encoder layers, token-level cross-attention module, and pretraining heads. 
% \noindent Details about the full architecture are as follows: 
% \subsection{Input Embeddings}
% \noindent \textbf{Text Embedding}
% Input text is a sequence of word tokens from a vocabulary $V$ appended with special tokens $[CLS]$ and $[SEP]$ at the start and end of the word tokens. A special token $MASK$ is used to randomly mask out word tokens with a small probability. We follow the preprocessing from BERT \cite{devlin2018bert} to process the text and use the same embedding layer for positional encoding of the text input.

% \noindent \textbf{Image Embedding}
% Our model takes convolutional features as image embeddings.
% % : spatial features or image patches. \textbf{Spatial Features:}
% The convolutional features $X \in \mathbb{R}^{h \times w \times d1}$ are extracted from a pre-trained ResNet model, $h, w$ is the feature height and width, and $d1$ is the extracted features dimension. A 2D convolutional layer then yields an embedding $X'$ of size $ \mathbb{R}^{h \times w \times d}$, where $d$ is the model dimension size. These input embeddings are used to produce capsule encodings $X_c$ as explained in section \ref{capsule_enc}.

\subsection{Capsule Networks} \label{sec:capsulenetwork}
% Capsule Networks (CapsNets) aim to explicitly model the part whole hierarchy relationship of entities within an image by learning their sub components, instantiation properties and the extent of their presence in the scene. CapsNets have shown state-of-the-art results in multiple vision fields, such as image interpretability, segmentation\cite{lalonde2018capsules}, 3D point clouds, and medical imaging\cite{lalonde2018capsules}.
% The promising performance of capsules can be attributed to their ability of learning part whole relationships for object entities via routing-by-agreement \cite{sabour2017dynamic} between different capsule layers. CapsNets are distinguished from CNNs as they are capable of learning view-point invariant representations of object entities. However, the idea of capsules has gained limited attention since it's introduction by Hinton et. al \cite{hinton2011transforming, sabour2017dynamic}. 
Convincing amount of evidence exist in psychology that humans parse visual scenes into part-whole hierarchies by modelling the viewpoint-invariant spatial relationship between a part and a whole~\cite{hinton1979some,kahneman1992reviewing}
% , as the coordinate transformation between the intrinsic coordinate frames assigned to them. 
Neural Networks can profit from understanding images in the same way humans do to become more transparent and interpretable. 
% if we try to make them understand images in the same way humans do. 
However, standard NNs lack this ability to dynamically represent a distinct part-whole hierarchy tree structure for each image\cite{hinton2021represent}.
% However, this is difficult for standard NNs because they cannot dynamically represent a different part-whole hierarchy tree structure for each image\cite{hinton2021represent}.

This inability motivated the introduction of a type of model called Capsule Networks \cite{hinton2011transforming} which was later formalized in \cite{sabour2017dynamic}. A Capsule Network is a neural network that is designed to model part-whole hierarchical relationships more explicitly than Convolutional Neural Networks (CNNs), by using groups of neurons to encode entities and learning the relationships between these entities. The promising performance of capsules can be attributed to their ability to learn part-whole relationships for object entities via routing-by-agreement \cite{sabour2017dynamic} between different capsule layers. A capsule is represented by a group of neurons; each capsule layer is comprised of multiple capsules and multiple capsules layers can be stacked together. Capsule routing is a non-linear, iterative and clustering-like process that occurs between adjacent capsule layers, dynamically assigning \textit{part} capsules $i$ in layer $\ell$ to \textit{object} capsules $j$ in layer $\ell+1$ , by iteratively calibrating the routing coefficients $\boldsymbol{\gamma}$~\cite{capsules_survey}. 
% Capsule network was proposed by Hinton et. al \cite{hinton2011transforming} to learn view-equivariant feature vectors from images. Sabour et. al \cite{hinton2018matrix} extended the idea further and use a routing-by-agreement to classify and segment multiple digits in images. Various works are proposed since then to improve routing mechanism and applying capsules to different problems including object detection and segmentation. 
Unlike most previous works which use a loss over object classes to learn a set of capsule classes, we do not have any object level supervision available for capsules, but instead combine the power of transformers and capsules by interleaving capsules as intermediate layers within the transformer and use VQA supervision to model visual entities as capsules.

\subsection{Capsule Encodings} \label{capsule_enc} 
% Capsules are known to learn objectness  of entities by learning their instantiation properties, and measure the object presence in the image \cite{sabour2017dynamic}. A capsule layer is composed of multiple capsules. 
% and multiple capsule layers can be stacked together. 
% The capsules in the lower layer use a voting mechanism to agree or disagree on the presence of a higher layer capsule known as routing-by-agreement.
We use matrix capsules \cite{hinton2018matrix} as follows:
% use EM-Routing for routing between capsule layers. 
given an image embedding $X'  \in \mathbb{R}^{h \times w \times d}$, matrix capsules $X_{c} \in \mathbb{R}^{h \times w \times d_c}$, as shown in Figure~\ref{fig:caps-arch}(a), are obtained as follows: The image embedding $X'$ is input to a convolutional layer producing primary capsules $X_p$ where each capsule has a pose matrix of size $K \times K$ and an activation weight. The primary capsule layer outputs $C_p$ number of capsules for each spatial location. 
% ($C_p$ is number of capsules in primary capsule layer).conda remove --name myenv --all
The output dimensions for poses is $\mathbb{R}^{h \times w \times C_p \times K \times K}$ and for activation is $\mathbb{R}^{h \times w \times C_p \times 1}$. To treat each capsule as a separate entity, the pose matrix and activation is grouped together for each capsule. Hence, the primary capsules $X_p$ have the dimensions $\mathbb{R}^{h \times w \times d_p}$ where $ d_p = C_p \times (K \times K + 1)$. The primary capsules are then passed through an EM-Routing layer to vote for capsules in the next layer. Assuming we have $C_v$ number of capsules in the next layer, the routing yields capsule encodings $X_{c}$ where $X_{c} \in \mathbb{R}^{h \times w \times d_c}$, $ d_c = C_v \times (K \times K + 1)$. We use an equal number of capsules in both layers i.e., $C=C_p=C_v$.
Our system employs the capsule representation $X_c$ as visual embeddings.
% We obtain the final capsule encodings $X_{c} \in \mathbb{R}^{h \times w \times d_c}$, where $d_c$ is the output dimension for capsules and denote $C$ capsules each comprised of pose matrix $K \times K $ and an activation score. ($d_c = C \times (K \times K + 1)$, $K$ is the pose-matrix size, 1 is the activation dimension). 

Since transformers take a sequence of tokens as input, we flatten the capsule embeddings across spatial dimension to get a sequence of visual tokens of length $h*w$, where each visual token is denoted by $x_j \in \mathbb{R}^{d_c}$ for $j \in {1,2,...,hw}$.
A special trainable token $[IMG]$ is then concatenated to these tokens to form the final set of visual tokens $\{[IMG], x_1, x_2, ..., x_{hw}\}$. A learnable position embedding is added to these visual tokens to keep the notion of spatial position in the sequence. Each of the visual tokens except $[IMG]$ is represented by $C$ capsules.

\subsection{Text-guided Capsule Selection} \label{masked_capsules} As the language encoder is attending different words at each layer, we select visual capsules based on the text representation at each visual encoder layer. Let $h^{i}_{cls}$ be the feature output corresponding to $[CLS]$ token from the $i^{th}$ text encoder layer; we take the feature output $h^{1}_{cls}$ against $[CLS]$ token from first text encoder layer and input to a fully connected layer $\phi$. The output is $C$ logits followed by a softmax function to learn presence probability $m^1 \in \mathbb{R}^C$ of attended words at layer 1. This mask is applied to $X_c$ to select the respective capsules and mask out the rest resulting in the masked capsule representation $X^1_{mc}$. 

 \begin{equation} \label{eq:1}
 m^1  = softmax(\phi(h^{1}_{cls})).
 \end{equation}
 
 \begin{equation} \label{eq:2}
    X^1_{mc} = m^1 \odot X_c 
\end{equation}

The masking is only applied to the visual tokens $x_j$ without affecting $[IMG]$ token.  
\\

\subsection{Text-based Residual Connection} \label{masked_residual} To keep the capsule representation between intermediate layers, we add capsules via a residual connection to the inputs of each intermediate visual encoder layer. The input capsules to the intermediate layer are also selected based on the intermediate features output from text encoder.  Let $m^i$ be the probability mask for attended words in the text feature output $h^i_{cls}$ from the $i^{th}$ layer:
\begin{equation} \label{eq:3}
    m^i  = softmax(\phi(h^{i}_{cls})), \forall i \in  \{1,2,...,L\},
\end{equation}

\noindent and $x^i_{j_{mc}}$ denotes the $j^{th}$ token with visual capsules selected using mask $m^i$.

\begin{equation} \label{eq:4}
    x^i_{j_{mc}} = m^i \odot X_c,
\end{equation}

 The $i^{th}$ visual encoder layer takes features from $(i-1)^{th}$ layer to produce features $h^i_{v_j}$ for $j^{th}$ position. Let $f^i_v$ be the $i^{th}$ layer in the visual encoder. The output and input follow the notation below:
\begin{equation} \label{eq:5}
    h^i_{v_j} = f^i_v (h^{i-1}_{v_j}).
\end{equation}

To keep reasoning flowing from text to image, we propose to add the residual connection from visual capsules for $j^{th}$ token by adding $x^i_{j_{mc}}$ to the input of $i^{th}$ encoder layer. However, there is a dimension mismatch between $x^i_{j_{mc}} \in \mathbb{R}^{d_c}$ and $h^{i-1}_{v_j} \in \mathbb{R}^d$. We upsample $x^i_{j_{mc}}$ to dimension size $d$ using a fully connected layer $\sigma$ and get the upsampled capsule-based features $\hat{x}^i_{j_{mc}} \in \mathbb{R}^d$. The input to the visual encoder layer will be as follows:

\begin{equation} \label{eq:6}
    \hat{h}^{i-1}_{v_j} = f^i_v (h^{i-1}_{v_j} + \hat{x}^i_{j_{mc}}).
\end{equation}

The output feature sequences from both encoders are then input to our cross attentional module which allows token-level attention between the two modalities. The aggregated feature outputs corresponding to $[CLS]$ and $[IMG]$ tokens after cross attention are input to a feature pooling layer followed by respective classifiers for pretraining and downstream tasks. 
We discuss the implementation about modality-specific encoders, feature pooling, and cross attention in detail in the supplementary.

\subsection{Training} \label{pretraining}
To perform well, transformers require pretraining on large-scale datasets before finetuning for the downstream tasks, i.e., GQA and VQA-HAT in our case. Therefore, we first pretrain our capsules-transformer backbone on three pretraining tasks: image-text matching (ITM), masked language modeling (MLM), and visual question answering (VQA). 
%Our initial experiments indicated that end-to-end training hurts the grounding ability. Hence 
The system is pre-trained in two stages: first, we do joint training of modality-specific encoders only to learn text-guided capsules representation; the representation learned in encoders is kept fixed during the second stage of pre-training where we add cross-attentional blocks on top of the modality encoders allowing token-level interaction between text features and visual features. While the first stage of pretraining uses pooled features from text and from visual encoders, the second stage pools features after cross attention: therefore, the second stage pre-training tasks uses cross-modal inputs as language and image features. For details about pretraining tasks in context of our method, refer to section 1.2 in supplementary. 
We finally finetune the pretrained capsules-transformer backbone to solve VQA as our downstream task.

%----------------------------------------------------------------------
\section{Experiments and Results}
\vspace{-2pt}

\subsection{Datasets}\label{VQADownstreamTask}
\noindent \textbf{Pre-training.}
We use MSCOCO \cite{lin2014microsoft} and Visual Genome \cite{krishna2017visual} for pretraining our backbone. We use the same data as \cite{tan2019lxmert} which also include MSCOCO-based VQA datasets: Visual7W, VQAv2.0, and GQA. However, we exclude the GQA validation set from pretraining and finetuning as we evaluate grounding on this set (scene graphs for GQA test and testdev are not publicly available). We use train sets of MSCOCO and VG with $\sim$7.5M sentence-image pairs for pretraining. MSCOCO val set is used for validating pretraining tasks. %For GQA, we follow the same pretraining split as \cite{khan2021reason}.

\noindent \textbf{Downstream.}\label{VQADownstreamTask1}
 We consider two datasets for the downstream task, GQA\cite{hudson2019gqa} and VQA-HAT\cite{vqahat}. 
 
\noindent \textbf{GQA} poses visual reasoning in the form of compositional question answering. It requires multihop reasoning to answer the question, so GQA is a special case of VQA. GQA is more diverse than VQA2.0 \cite{goyal2017making} in terms of coverage of relational, spatial, and multihop reasoning questions. It has 22M QA pairs with 113K images. GQA provides ground truth boxes for question and answer objects making it a suitable test bed for our task. 

\noindent \textbf{VQA-HAT dataset} provides human attention maps for VQA task. This dataset is based on VQA v1.0\cite{antol2015vqa} dataset and provides 1374 QA pairs with 488 images in validation set. 
To evaluate on this dataset, we train our system on VQA v1.0.
% and evaluate on VQA-HAT validation set.  
%To evaluate on VQA-HAT, we train on VQA v1.0 \cite{antol2015vqa} dataset.
% We use train and validation set (121,512 questions) for finetuning. 
The answer vocabulary of VQA train set has a long tail distribution. We follow previous works \cite{antol2015vqa,vqahat} and use 1000 most frequent answers. We first combine training (248,349 QA pairs) and validation data (121,512 QA pairs) to get a total of 368487 QA pairs. We then filter out the questions with out-of-vocabulary answers
% (answer vocab size is kept 1K)
resulting in 318827 QA pairs. 
%We separate out 10K QA pairs from the training set (after above mentioned question filtering) and use it as a validation set to pick our best model. We therefore use 308K QA pairs from VQA v1.0 train and val set for finetuning our pretrained backbone with 16 capsules. The learning parameters used for this training are lr=4e-5, batch size=64, with bert optimizer and trained for 20 epochs. The best model on validation set is used for evaluation. \hkc{goes to supplementary ... ?}

% We obtain mean
% rank-correlation score with avg. human attention map of \textbf{0.479} vs. HieCoAtt-Q (0.264).
% \paragraph{Pre-training details}
% We pretrain our system in two stages. In first stage, the visual encoder and language encoder are trained with all pretraining tasks. In second stage, a multilayer cross-attentional transformer encoder is added on top of both encoders. We freeze the weights for both modality-specific encoders, and only train the cross-attention module allowing direct interaction between vision and text. We find that using 2-stage pretraining gives us best grounding results.

\subsection{Evaluation Metrics}

For GQA, VQA accuracy is reported for task accuracy. For grounding task on transformers, we take attention scores from [IMG] token to image from the last cross-attentional layer for all heads. 
% Each head is producing an attention map. 
Answer grounding performance is evaluated in terms of the following: \textbf{Overlap}-- overlap between the ground truth bounding box for answer object and the detected attention region is reported in terms of precision (P), recall (R), and F1-score (F1); \textbf{IOU}-- intersection over union (IOU) between the ground truth object and detected region is reported in terms of P, R, and F1-score.
% \textbf{Grounding}--the grounding metric proposed by \cite{hudson2019gqa} as the average sum of attention over ground truth region(s) r for all data samples;
\textbf{Pointing Game}-- proposed by \cite{zhang2018top-pgacc} is a metric for weakly-supervised visual grounding methods. For pointing game, we consider the point detected from each head as a part of distribution, and perform k-means clustering (k=1) on those points. The cluster center is considered as the detected point from the system and used for evaluating accuracy. For VQA-HAT, we report \textbf{mean rank correlation} between system generated attention and human attention maps to compare with previous methods.

\subsection{Implementation details.}
We use $L=5$ layers in both text and image encoders, and $2$ layers in cross-attention module. The transformer encoder layers have the same configuration as BERT with 12 heads and feature dimension $d=768$. A batch size of 1024 with learning rate $lr=1e-4$ is used for pretraining. First stage pretraining is done for 20 epochs and further trained for 10-15 epochs during second stage. We use Imagenet pre-trained ResNet model to extract features of dimensions $7\times7\times2048$. For finetuning on GQA, we use batch size=32 with $lr=1e-5$ and 5-10 training epochs. For VQA-HAT, we use batch size=64 with $lr=4e-5$ trained for 20 epochs. 
%No training parameters search is done at any stage of training. \\

To evaluate the grounding results, we follow the best practice of DINO~\cite{dino} and take the attention map from the last cross-attentional layer. To compute overlap and IOU for GQA, we threshold over the attention map with an attention threshold of 0.5 to get high attention regions. Each connected region is considered a detection. For pointing game, we find the single cluster center over maximum attention points from all heads and use it for evaluation. We ignore the test samples with empty ground truth maps for pointing game since there is no ground truth bounding box to check for a hit or a miss. For the VQA-HAT evaluation, we follow~\cite{vqahat} and use mean rank correlation between the generated attention maps and the ground truth.

\begin{table}[t]
\small
\renewcommand{\arraystretch}{.9}
  \centering \setlength{\tabcolsep}{.9\tabcolsep}   
    \begin{tabular}{lcc}
       \toprule
                % \multicolumn{1}{c}{\textbf{Method}} & \multicolumn{3}{c}{}  \\
                % \cmidrule(lr){2-4}

 Method  & Layer &  Pointing Game Acc. \\
%  \multicolumn{1}{c}{\textbf{Method}}   & Layer &  $\tau=0$ & $\tau=5$ & $\tau=10$ & $\tau=15$\\
   \midrule
   Random & - &  18.80 \\
     Center & - &  33.30\\
     \midrule  
    MAC \cite{hudson2018compositional} & mean &  8.90  \\
                            % & object &   &  &  &  &  &  \\
                            % \midrule
    MAC-Caps \cite{khan2021reason} &  mean  &  28.46  \\
     LXMERT \cite{tan2019lxmert} &  last   &  29.00 \\
    %  ALBEF \cite{ALBEF} -Grounding & last &  29.11 \\
    %  ALBEF \cite{ALBEF} -VQA & last &  30.66  \\
     ALBEF \cite{ALBEF}-GC & last & 32.13\\
     ALBEF \cite{ALBEF}-ATN & last & 32.11\\
     ViLT \cite{vilt} & last & 11.99  \\
     \midrule
    %  Ours - no skip & last & 29.81 \\
    %  Ours - ViT  & last &  27.06 \\
    Ours-no-init (C=16)  & last &  \textbf{34.59}\\
        Ours-no-init (C=32)  & last &  \textbf{34.43}\\
        Ours-nogqa  (C=32)  & last &  \textbf{37.04} \\
    %   Ours - DINO  &  & &\\
   
	\bottomrule
    \end{tabular}
 \vspace{2pt}
    \caption{\footnotesize Pointing game accuracy for GQA. For MAC and MAC-Caps, mean attention maps over reasoning steps are used. For transformer-based methods, maximum attention points from all heads are used for clustering. The cluster center is then used for the pointing game evaluation. For ALBEF, GC=GradCAM output, ATN=attention output. 
    % Ours-no skip=ours without residual connections, Ours-ViT=visual encoder is initialized from pre-trained ViT weights, 
    Ours-no-init is the full model with residual connections and trained from scratch, Ours-nogqa uses no GQA samples at pretraining stage. Numbers are reported in percentages.}
    \label{tab:GQA_grounding_MAC_PG}
\vspace{-20pt}
\end{table} 

%----------------------------------------------------------------------

%----------------------------------------------------------------------

\begin{table*}[t]
\scriptsize
\renewcommand{\arraystretch}{.9}
  \centering \setlength{\tabcolsep}{.4\tabcolsep}   
    \begin{tabular}{lcccccccccccc}
       \toprule
                % \multicolumn{1}{c}{\textbf{Method}} & \multicolumn{3}{c}{}  \\
                % \cmidrule(lr){2-4}

    & & & &  & & \multicolumn{3}{c}{Overlap} & & \multicolumn{3}{c}{IOU}   \\
    \cmidrule{7-9} \cmidrule{11-13} 
   \multicolumn{1}{c}{\textbf{Method}} & Obj. & Backbone & Pre-training & Layer & Acc. & P & R & F1 & & P & R & F1 \\
      
        \midrule

    MAC \cite{hudson2018compositional} & A & ResNet & - & last & 57.09 & 5.05 & 24.44 & 8.37 & & 0.76 & 3.70  & 1.27 \\
                            % & object &   &  &  &  &  &  \\
                            % \midrule
    MAC-Caps \cite{khan2021reason} & A & ResNet & - & last & 55.13 & 5.46 & 27.9 & 9.13 & & 0.97 & 4.94 & 1.62\\
   
    \midrule
    
     LXMERT-patches & A & Faster RCNN & MSCO,VG & last  & 48.65 & 7.13 & 64.21 & 12.83 && 0.95 & 8.66 & 1.71 \\
    %   LXMERT-patches \\
    %   LXMERT-grid & & 64.08 & 15.52 & 81.19 & 26.06 && 2.23 & 11.74 & 3.74\\ 
    %   Ours-X3-patches  & last & - &  6.60 & 78.23 & 12.18 && 0.76 & 9.30 & 1.40\\
      \midrule 
  
      %ALBEF \cite{ALBEF} -Grounding (A)  & last & - & 0.01 & 0.44 & 0.02 && 0.01 & 0.27 & 0.02\\
      %ALBEF \cite{ALBEF} -Grounding (Q)  & last & - &  0.02 & 0.20 & 0.03 && 0.01 & 0.10 & 0.02\\
      %ALBEF \cite{ALBEF} -VQA (A) & last & - & 0.01 & 0.06 & 0.02 && 0.01 & 0.03 & 0.01\\
    %   ALBEF \cite{ALBEF}-VQA -GC  & last & - & 2.85 & 99.92 & 5.54   && 0.37 & 13.31 & 0.71\\
    %   ALBEF \cite{ALBEF}-VQA -ATN  & last & - &  &  &   &&  &  & \\
    %   ALBEF \cite{ALBEF}-VQA -GC (A) & 8 & - & 2.94 & 99.92 & 5.72  && 0.38 & 13.65 & 0.74\\
    %   ALBEF \cite{ALBEF}-VQA -ATN (A) & 8 & - & 3.72 & 99.92 & 7.17  &&  0.46 & 12.39 & 0.88\\
      ALBEF \cite{ALBEF}-GC & A & ViT+BERT & & last & 64.16 & 6.94 & 99.92 & 12.98 && 0.89 & 13.43 & 1.67\\
    ALBEF \cite{ALBEF}-ATN & A & ViT+BERT & MSCO,VG, & last & 64.20 & 5.13 & 99.92 & 9.75 && 0.64 & 12.98 & 1.21\\
      ALBEF \cite{ALBEF}-GC & A & ViT+BERT & SBU,GCC & 8 & 64.20 & 4.41 & 99.92 & 8.44 && 0.54 & 12.85 & 1.04 \\
      ALBEF \cite{ALBEF}-ATN & A & ViT+BERT & & 8 & 64.20 & 4.82 & 99.92 & 9.19 && 0.60 & 12.88 & 1.14\\
      ViLT \cite{pmlr-v139-kim21k} & A & ViT & MSCO,VG, & last-cos & 66.33 & 0.34 & 6.13 & 0.65 && 0.04 & 0.63 & 0.07\\
        ViLT \cite{pmlr-v139-kim21k} & A & ViT & SBU,GCC & last-ATN& 66.33 & 0.28 & 4.10 & 0.53 && 0.08 & 1.20 & 0.15\\
    %   SOHO \cite{soho} \\
      \midrule
     Ours (C=16) & A & ResNet & MSCO,VG & last  & 56.65 & \textbf{14.53} & 85.47 & \textbf{24.84} && \textbf{2.30} & \textbf{13.61} & \textbf{3.94}  \\
     
     \midrule
     \midrule
      MAC \cite{hudson2018compositional} & Q & ResNet & - & last & 57.09 & 10.79 & 16.38 & 13.01 && 1.39 & 2.09 & 1.67  \\
                            % & object &   &  &  &  &  &  \\
                            % \midrule
    MAC-Caps \cite{khan2021reason} & Q & ResNet & - & last & 55.13 & 17.39 & 28.10 & 21.49 && 1.87 & 2.96 & 2.29\\
   
    \midrule
    
     LXMERT-patches & Q & Faster RCNN & MSCO,VG & last & 48.65 & 32.46 & 64.02 & 43.08 && 3.48 & 6.87 & 4.62 \\
 
      \midrule

    %   ALBEF \cite{ALBEF}-VQA -GC  & last & - & 21.99 & \textbf{99.91} & 36.05  && 2.24 & \textbf{23.02} & 4.08\\
    %   ALBEF \cite{ALBEF}-VQA -ATN  & last & - &  &  &   &&  &  & \\
    %   ALBEF \cite{ALBEF}-VQA -GC (A) & 8 & - & 9.33 & 99.90 & 17.06 && 0.81 & 8.98 & 1.48 \\
    %   ALBEF \cite{ALBEF}-VQA -ATN (A) & 8 & - & 11.88 & 99.89 &  21.23 && 0.96  & 8.07 & 1.71 \\
      ALBEF \cite{ALBEF}-GC & Q & ViT+BERT & & last & 64.20 & 22.15 & 99.90 & 36.26 && 1.96 & 9.22 & 3.24\\
      ALBEF \cite{ALBEF}-ATN & Q & ViT+BERT & MSCO,VG, & last & 64.20 & 16.50 & 99.90 & 28.33 && 1.40 & 8.90 & 2.43\\
      ALBEF \cite{ALBEF}-GC & Q & ViT+BERT & SBU,GCC & 8 & 64.20 & 14.21 & 99.90 & 24.88 && 1.19 & 8.71 & 2.09\\
      ALBEF \cite{ALBEF}-ATN & Q & ViT+BERT & & 8 & 64.20 & 15.51 & 99.90 & 26.85 && 1.31 & 8.77 & 2.27\\
      ViLT \cite{pmlr-v139-kim21k} & Q & ViT & MSCO,VG, & last-cos& 66.33 & 1.02 & 5.64 & 1.73 && 0.10 & 0.54 & 0.17\\
      ViLT \cite{pmlr-v139-kim21k} & Q & ViT & SBU,GCC & last-ATN& 66.33 & 0.34 & 1.56 & 0.56 && 0.08 & 0.38 & 0.14 \\
    %   SOHO \cite{soho} \\
      \midrule
     Ours (C=16)  & Q & ResNet & MSCO,VG & last & 56.65 & \textbf{47.03} & 81.67 & \textbf{59.69} && \textbf{4.72} & 8.08 & \textbf{5.96}\\
     
    %  Ours w/skip (C=48,X1)  & last  & 53.65 & 10.28 & 68.94 & 17.89 && 1.59 & 10.73 & 2.78   \\

      %Ours-X1-nogqa-2st  & last (mean) & 55.41 & 9.89 & 71.77 & 17.38 && 1.59 & 11.55 & 2.79\\
      
      %Ours-X1-ViT-BERT (C=32) & & \textbf{58.86} &  8.54 & 60.15 & 14.95 && 1.22 & 8.63 & 2.13\\
      %Ours-X1-ViT-BERT (C=24) & & \textbf{56.40} &  - & - & - && - & - & -\\
      %Ours-X2-ViT-BERT (C=32) & & \textbf{58.83} &  - & - & - && - & - & -\\
      %Ours-X3-ViT-BERT (C=32) & & \textbf{56.41} &  - & - & - && - & - & -\\
    %   \hline
      %Ours-X1-no-init (C=24) & & 56.26 & 10.90 & 74.03 & 19.00 && 1.54 & 10.56 & 2.69 \\
      %Ours-X1-no-init (C=32) & & 55.41 & 10.09 & 71.95 & 17.70 && 1.41 & 10.09 & 2.47 \\
      %Ours-X1-no-init (C=48) & & 53.65 & 10.28 & 68.94 & 17.89 && 1.59 & 10.73 & 2.78 \\
      %Ours-X2-no-init & & 55.71 \\
    
\bottomrule

    \end{tabular}
 \vspace{5pt}
    \caption{\small Results on GQA validation set (for last layer). All methods are evaluated for weak VQA grounding task. For transformer-based models, attention was averaged over all heads. Results are based on grounding of objects referenced in the answer (A) and the question (Q). C=no.of capsules, we report results from our best model with C=16. Refer to table \ref{tab:ablations_mean} for more variants. For ViLT, we obtain results using cosine similarity (cos.) between text and image features as proposed by the authors as well as from raw attention scores (ATN). For ALBEF, GC is the gradcam output used for evaluation, ATN is the attention output. ALBEF uses layer 8 as grounding layer, we also report grounding performance on this layer. Our method outperforms all baselines for overlap F1-score and IOU F1-score. See section \ref{results} for more details.
    Numbers are reported in percentages.}
    \label{tab:GQA_grounding_MAC_last_A}
    \vspace{-3mm}
% \vspace{-15pt}
\end{table*}

%----------------------------------------------------------------------

\subsection{Comparison to State-of-the-Art} \label{results}

We compare the performance of our method to other best-performing methods in the field of weakly supervised VQA grounding and VQA in general, namely MAC \cite{hudson2018compositional} and MAC-Caps \cite{khan2021reason} as representation of visual reasoning architectures, and LXMERT \cite{tan2019lxmert}, ViLT \cite{vilt}, and ALBEF \cite{ALBEF} as state-of-the-art transformer architectures without object features.
%For MAC and MAC-Caps, we compare to last attention maps from following the same evaluation protocol from \cite{khan2021reason}.

For LXMERT, we take the provided backbone pre-trained on object features and finetune it using image patches of size  $32 \times 32 \times 3$ on GQA. 
In case of ViLT, we use the provided pre-trained backbone and finetune it on GQA. Following ViLT, we generate a heat map using the cosine similarity between image and word tokens evaluating the similarity scores as well as raw attention for grounding performance for all three metrics.
For ALBEF we report results on last layer as well as on layer 8 which is specialized for grounding~\cite{ALBEF} using the visualizations from gradcam as well as raw attention maps. 

%\noindent \textbf{MAC and MAC-Caps:}
% MAC-Caps \cite{khan2021reason} is the state-of-the-art for answer grounding on GQA. 
\noindent \textbf{GQA} We first look at the results of our evaluation on GQA, considering pointing game accuracy in Table~\ref{tab:GQA_grounding_MAC_PG} and for overlap and IOU in  Table~\ref{tab:GQA_grounding_MAC_last_A}. Our method outperforms both MAC and MAC-Caps for answer grounding on the last attention map. We achieve an absolute gain of 16.47\% (overlap F1-score) and 2.67\% increase in IOU F1-score, and an improvement of 25.69\% in pointing game accuracy for MAC. When compared with MAC-Caps, our best method (C=16, no-init) improves overlap F1-score by 15.71\% $\uparrow$, IOU F1-score by 2.32\% $\uparrow$, and pointing game accuracy by 6.13\% $\uparrow$. 
% Ours results even outperform MAC-Caps' best step results by 4.77\% for overlap F1-score, and a slight improvement in IOU F1-score (0.06\% $\uparrow$)  (see table \ref{tab:GQA_grounding_MAC_PG} and table \ref{tab:GQA_grounding_MAC_last_A}).
Similar performance gain is observed for question grounding with an improvement of 38.2\% $\uparrow$ for overlap F1-score and 3.67\% $\uparrow$ gain for IOU F1-score.
% For question grounding i.e. to evaluate grounding for objects referenced in the question, we achieve 59.69\% in terms of overlap F1-score, () and IOU F1-score of 5.96\% () compared to MAC-Capsules (see table \ref{tab:GQA_grounding_MAC_last_Q}). We also outperform MAC and MAC-Capsules in terms of VQA accuracy. 

%\noindent \textbf{LXMERT:} 
To evaluate LXMERT finetuned on image patches (LXMERT-patches), we take the attention score maps from the last cross modality layer. 
% Attention maps is averaged over all attention heads. 
% LXMERT obtain the second highest scores in overlap F1-score (12.83\%) and IOU F1-score (1.71\%),
We improve over LXMERT by 12.01\% $\uparrow$ absolute points w.r.t overlap F1-score, 2.23\% $\uparrow$ w.r.t. IOU F1-score and 5.43\% $\uparrow$ gain in pointing game accuracy. For question grounding, LXMERT achieves an overlap F1-score: 43.08\% (vs. ours 59.69\%) and IOU F1-score of 4.62\% (vs. ours 5.96\%).

%\noindent \textbf{ViLT:} 
ViLT outperforms all methods in terms of VQA accuracy of 66.33\%. However, on the grounding task, it demonstrates the lowest performance for all metrics (table \ref{tab:GQA_grounding_MAC_PG}: row 7, table \ref{tab:GQA_grounding_MAC_last_A}: rows 8-9). Similar behavior is observed for the question grounding task. % We evaluate VilT with the heatmaps produced using post processing proposed by authors as well as with attention scores for image patches for a fair comparison with rest of the systems.

%\noindent \textbf{ALBEF:} 
% We compare our method with two flavors of ALBEF. First, we evaluate their finetuned model on VQAv2.0 \cite{balanced_vqa_v2} provided by the author for grounding task. 
ALBEF produces visualization using GradCAM. 
% since we evaluate our method and the baselines using attention maps, 
We compare with ALBEF using both GradCAM output and attention maps. ALBEF has a very high recall (R) both in terms of overlap and IOU. However, it lacks in precision (P) leading to lower F1-scores for both metrics. Our best model outperforms ALBEF-VQA by a significant margin on both answer grounding and question grounding. 

\begin{table}[t]
   \centering \small
    \renewcommand{\arraystretch}{.6}
  \centering \setlength{\tabcolsep}{.5\tabcolsep}
    \begin{tabular}{lc}
       \toprule
   \multicolumn{1}{c}{\textbf{Method}} &  \textbf{Mean Rank-Correlation}
    \\
    \midrule
    Random &   0.000 $\pm$ 0.001\\
    Human  &  0.623 $\pm$ 0.003 \\
    \midrule
    \textit{Unsupervised} \\ \\
    SAN  \cite{yang2016stacked} &  0.249 $\pm$ 0.004 \\
    HieCoAtt \cite{lu2016hierarchical} &  0.264 $\pm$ 0.004 \\
    Ours (C=16) &  \textbf{0.479 $\pm$ 0.0001} \\
    \midrule
    \textit{Supervised}\\ \\
    HAN \cite{qiao2018exploring} & 0.668 $\pm$ 0.001\\
      
    \hline
	\bottomrule
    \end{tabular}
 \vspace{5pt}
    \caption{ Results on VQA-HAT val dataset. \textit{Unsupervised}: no attention supervision, \textit{Supervised}: use attention refinement.
    }
    \vspace{-15pt}
    \label{tab:vqa-hat}

\end{table}

\begin{figure}[t]
\begin{center}

  \scriptsize
      \centering
  \includegraphics[width = 1\linewidth]{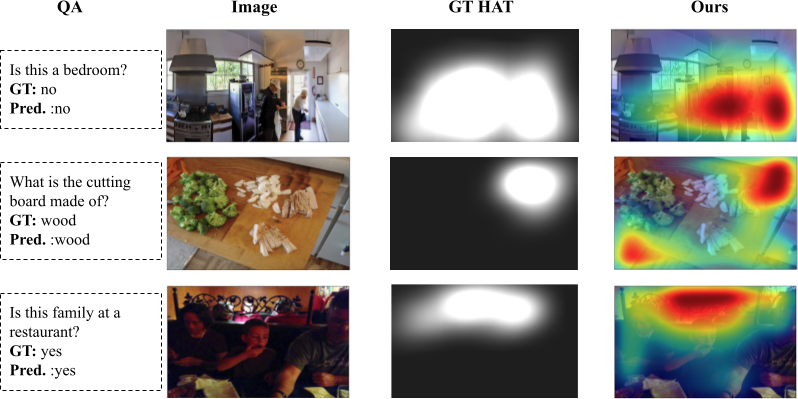}
  \captionof{figure}{\textbf{Success cases for VQA-HAT dataset.} VQA-HAT provides 3 human attention maps for each image. Here, we show the best matched ground truth map (GT HAT). }
   \vspace{-15pt}
\label{fig:vqa-hat-qual}
\end{center}
\end{figure}

\noindent \textbf{VQA-HAT} We further evaluate our system on the VQA-HAT dataset. To this end, we follow the protocol of VQA-HAT and resize the human attention maps and the output attention maps from our system to the common resolution of 14x14. 
We then rank both of them. VQA-HAT val set provides three human attention maps for each question. 
We compute the rank correlation of generated attention map with each human attention map and take the average score.
% We first scale the values between 0 and 1 by dividing with the maximum value. Then, we take the average of the provided 3 attention maps, rank the average map as well as our generated attention scores, and compute rank correlation between them.
Mean rank correlation score over all QA pairs is reported. 

We compare our approach on VQA-HAT with three different baselines: SAN~\cite{yang2016stacked} and HieCoAtt~\cite{lu2016hierarchical} as unsupervised bounding box free systems, and HAN~\cite{qiao2018exploring} which uses attention supervision during training. The evaluation is shown in table \ref{tab:vqa-hat}. It shows that the proposed system is able to significantly outperform both methods using VQA-only supervision. 

Without any attention supervision during training, we are able to narrow the gap between unsupervised methods and methods such as HAN, which uses human ground truth attention maps during training. 

Figure \ref{fig:vqa-hat-qual} shows success cases on VQA-HAT, comparing our generated attention result to the closest human attention map. 

%-----------------------------------------------------------
%  \vspace{-25pt}
\begin{table}[t]
\small
\renewcommand{\arraystretch}{.95}
  \centering \setlength{\tabcolsep}{.9\tabcolsep}   
    \begin{tabular}{lccccccccc}
       \toprule
                % \multicolumn{1}{c}{\textbf{Method}} & \multicolumn{3}{c}{}  \\
                % \cmidrule(lr){2-4}

      &   & \multicolumn{3}{c}{Overlap} && \multicolumn{3}{c}{IOU} & Pointing \\
    \cmidrule{3-5} \cmidrule{7-9} 
   \multicolumn{1}{c}{\textbf{Method}} & Acc.
     &  P & R & F1 & & P & R & F1 & Game\\
      
        \midrule 
    %  (1)  no caps.(X1)     & 62.43 & 11.88 & 65.42 & 20.11 && 1.77 & 9.80 & 3.00 & 34.22 \\
     % (2)  w/o mask       & 95.64 & - & - & - && - & - & - & -\\
     (1)  no skip(C=32) & 56.83~| & 11.06 & 77.60 & 19.37 &|& 1.39 & 9.85 & 2.43 & 29.81 \\ 
      
    %  (3)  e2e(C=32,X1)    & 57.02 & 10.36 & 64.28 & 17.84 && 1.33 & 8.30 & 2.29 & -\\
    %  (4)  2-stage(C=32,X1) & 55.41    & 10.09 & 71.95 & 17.70 && 1.41 & 10.09 & 2.47 & -\\
     %(5)  2-stage, freeze wts(X1)  & - & - & - & - && - & - & -& -\\
     (2)  w/skip (C=32)  & 55.41~| & 10.09 & 71.95 & 17.70 &|& 1.41 & 10.09 & 2.47 & 34.43\\
        \midrule 
        % (3)  w/skip (C=4)  & 55.15 &  &  &  &&  &  &  & \\
     (3)  w/skip (C=16)  & 56.65~| & 14.53 & 85.47 & 24.84 &|& 2.30 & 13.61 & 3.94 & 34.59\\
     (4)  w/skip (C=24)  & 56.26~| & 10.90 & 74.03 & 19.00 &|& 1.54 & 10.56 & 2.69 & 31.08\\
     (5)  w/skip (C=32)  & 55.41~| & 10.09 & 71.95 & 17.70 &|& 1.41 & 10.09 & 2.47 & 34.43\\
     (6)  w/skip (C=48)  & 53.65~| & 10.28 & 68.94 & 17.89 &|& 1.59 & 10.73 & 2.78 &  29.70  \\
    %  \midrule
     %(10)  w/skip (C=32,X1)  & 55.41 & 28.64 & 99.46 & 44.47 && 6.15 & 28.04 & 10.09 & 34.43\\
     %(11)  w/skip (C=32,X2)  & 55.71 & - & - & - && - & - & - & -\\
     %(12)  w/skip (C=32,X3)  & 55.78 & 35.22 & 99.99 & 52.09 && 6.98 & 27.06 & 11.09 \\
    %  (7)  w/skip (C=32,X1)  & 55.41 & 10.09 & 71.95 & 17.70 && 1.41 & 10.09 & 2.47 & 34.43\\
    %   Ours w/skip (C=32, X2) & last  & 55.78 & 	12.22 &	77.16 & 21.10 && 1.68 & 10.73 & \textbf{2.91} & -\\
     \midrule
     (7)  no-init (C=32)  & 55.41~| & 10.09 & 71.95 & 17.70 &|& 1.41 & 10.09 & 2.47 & 34.43\\
    %  (14) bert init (C=32, X1)  & - & - & - & - && - & - & - & -\\
     (8) vit-bert init (C=32) & 58.86~| & 11.11 & 74.67 & 19.34 &|& 1.55 & 10.44 & 2.69 & 27.06\\
    \hline
% 	\bottomrule
    \end{tabular}
 \vspace{3pt}
    \caption{\small Ablations over the design choices for the proposed architecture on GQA val set. Average attention over all heads in the last transformer layer is used to evaluate the grounding performance. We perform ablation study with C=32 caspules except rows 3-6 where we train the proposed architecture with varying number of capsules. Ablation (1) no skip is our system without residual connections, (2) w/skip is the full model. Rows 7 and 8 is the ablation over weight initialization from pre-trained transformers for vision (ViT) and text (BERT). 
    % The top 3 rows show results for models trained with 16 capsules.
    }
    % \vspace{-15pt}
    \label{tab:ablations_mean}
    \vspace{-1.0cm}

\end{table} 
%----------------------------------------------------------------------

\subsection{Ablations and Analyses} \label{ablations}

\noindent \textbf{Impact of Residual Connections.}
We compare our full system with an ablated variant without residual connections. We observe a drop in performance in terms of overlap, but a slight increase in terms of IOU. Without residual connections, pointing game accuracy is lower than with residual connections (4.62\% $\downarrow$). We conclude that using residual connections is beneficial for pointing game.  

\noindent \textbf{Number of capsules.}
We ablate our system with varying number of capsules. We train the proposed system with C=16, 24, 32, and 48 capsules. We observe that increase in number of capsules not only decreases VQA accuracy, but also hurts the overlap and IOU in terms of precision, recall and F1-score. Our best method therefore uses 16 capsules with residual connections and pre-trained from scratch. 

\noindent \textbf{ViT + BERT + Ours.} ViLT and ALBEF initialize their image and text encoders from ViT and/or BERT weights. 
% Similary, ALBEF uses ViT and BERT pretrained weights to initialize their vision and text transformers. 
Although our model is shallow than both models (5 layers in modality specific encoders compared to 12 layers in ViLT and ALBEF), we experimented to initialize our text encoder with BERT weights and image encoder with ViT weights from last 5 layers. We find a gain in VQA accuracy (58.86\% vs. 56.65\%) but it hurts the grounding performance. 

\begin{figure}[t]
\begin{center}
\includegraphics[width=\linewidth]{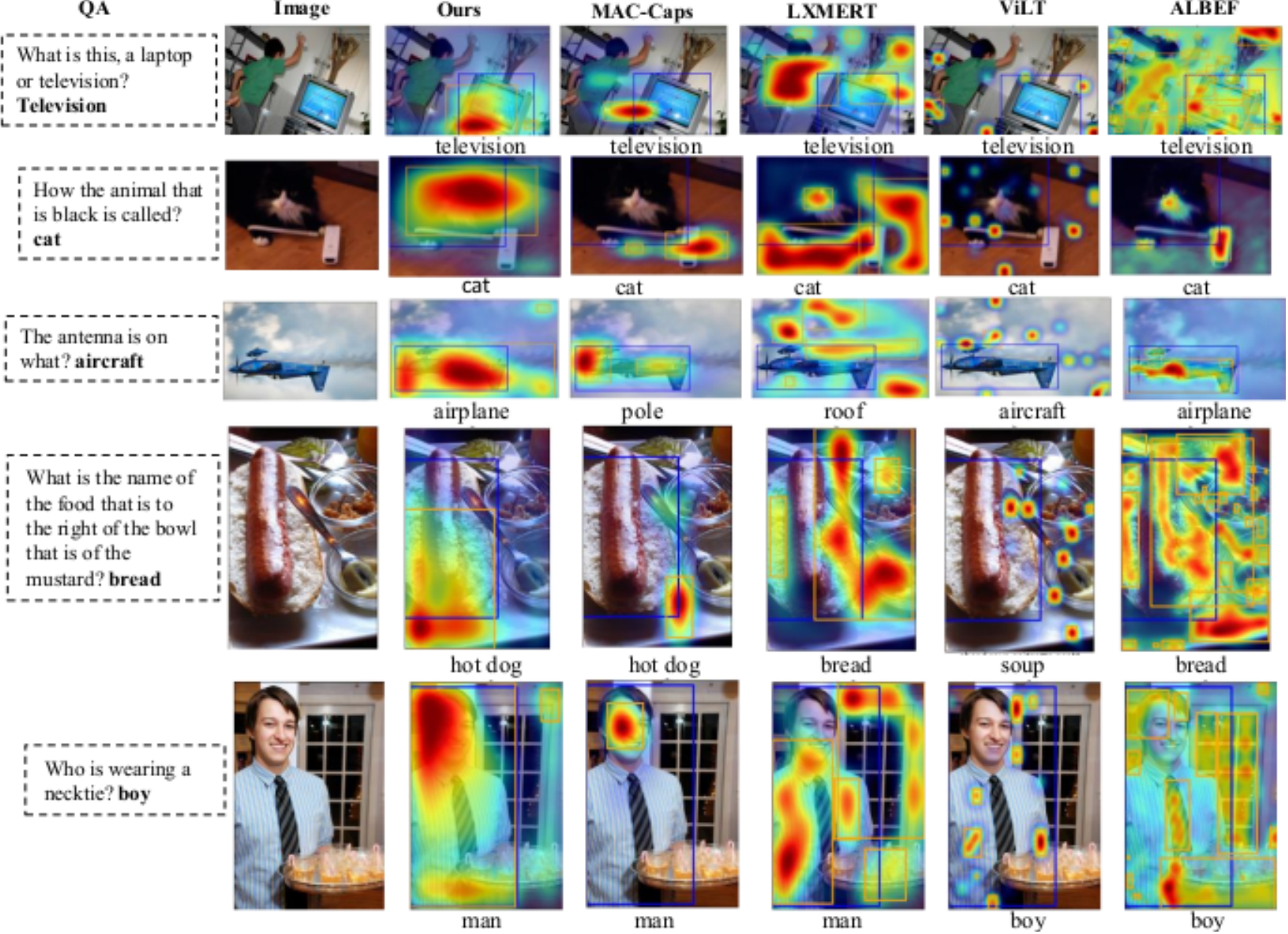}
\end{center}
\vspace{-20pt}
   \caption{\small \textbf{Qualitative comparison:} each row shows the input example, and the last layer's attention visualizations (averaged over all heads) with the predicted answer from all methods. Column 1 shows the question and ground truth answer, column 2 is the input image, column 3 shows the attention output from our method, column 4-7 are results from the baselines. Blue box is the ground truth bounding box for the answer object, orange boxes are the detected regions from each system. We can see that ours is attending relevant answer object with the plausible predicted answer even when the prediction mismatches with the ground truth answer (row 3-5). In row 4, the question is vague; therefore we can say, except LXMERT, all methods choose the correct answer. ALBEF has attention spread over all image which explains the high recall it achieves for overlap and IOU. Refer to section \ref{qualitative} for more details and discussion. Best viewed in color. 
   }
   \vspace{-15pt}
\label{fig:qualitative}
\end{figure}

\subsection{Qualitative Analysis} \label{qualitative}
In Figure \ref{fig:qualitative}, we show some qualitative comparison of our method with the baseline methods. 
% Visualization are obtained from mean attention map over all heads for last layer. 
For all examples including the ones where our system predicted the wrong answer, the grounding attention was correct (row 1,4 and 5). Also, the answers are plausible. For instance, in row 3, the correct answer is 'aircraft', and our method predicted it as 'airplane' with the correct localization. Overall, we notice that compared to our method, the baselines were either attending most of the image (ALBEF in rows 1,3, and 5 which explains the high recall in table \ref{tab:GQA_grounding_MAC_last_A}), or generate small attention maps (MAC-Caps, ViLT) or looking at the wrong part of the image (LXMERT). More examples and analysis are in the  supplementary document. 

\section{Conclusion}
% We evaluate grounding ability of the proposed method by taking the attention outputs from the last cross-attentional layer. However, there is a possibility that some intermediate layer does the better job at grounding such as ALBEF finds that layer 8 in their model is good at grounding. However, searching for the best layer is expensive in terms of time and computational cost. Nonetheless, our approach outperforms on grounding even when evaluated for the last layer only. Answer grounding is challenging because sometimes the answer word is not available in the input. An ideal VQA grounding system should perform well in terms of IOU (low IOU scores show the room for improvement on this metric). 
In this work, we show the trade-off between VQA accuracy and the grounding abilities of the existing SOTA transformer-based methods. We propose to use text-guided capsule representation in combination with transformer encoder layers. Our results demonstrate significant improvement over all baselines for all grounding metrics. Extensive experiments demonstrate the effectiveness of the proposed system over the baselines.

\clearpage

\bibliographystyle{splncs04}
\bibliography{egbib}
\appendix

\newpage
\section*{\centering Supplementary: Weakly Supervised Grounding for VQA in Vision-Language Transformers }

In this supplementary document, we further discuss about the proposed work as follows:
\begin{itemize}
    \item Additional implementation details (section \ref{extra-impl-details})
    \item Additional architecture details (section \ref{add-arch-details})
    \item Training objectives' details (section \ref{train-objectives})
    \item Grounding Performance Evaluation (section \ref{grounding-eval})
    \item Performance analysis w.r.t. varying detection threshold (section \ref{vary-threshold})
    \item Grounding accuracy w.r.t. each head (section \ref{vary-head})
    \item Results w.r.t. question type (section \ref{qtype})
    \item Entities represented by visual capsules (section \ref{capsule-vis})
     \item Training parameters in our method (section \ref{train-param})
     \item VQA accuracy comparison (section \ref{vqa-acc})
     \item Additional details about evaluation on VQA-HAT (section \ref{vqa-hat})
    \item Qualitative Results (section \ref{qual})
    % \begin{itemize}
    %     \item More examples for qualitative comparison with baselines (figures \ref{fig:qualitative2} and \ref{fig:qualitative3})
    %     \item More qualitative examples for our method (figures \ref{fig:qualitative4}, \ref{fig:qualitative5}, and \ref{fig:qualitative6})
    % \end{itemize}

    % \item Mean attention visualizations for image and text in each layer from the proposed model 
\end{itemize}

\begin{figure}[ht]
\centering
\begin{tabular}{c}
 {\includegraphics[width=0.7\linewidth]{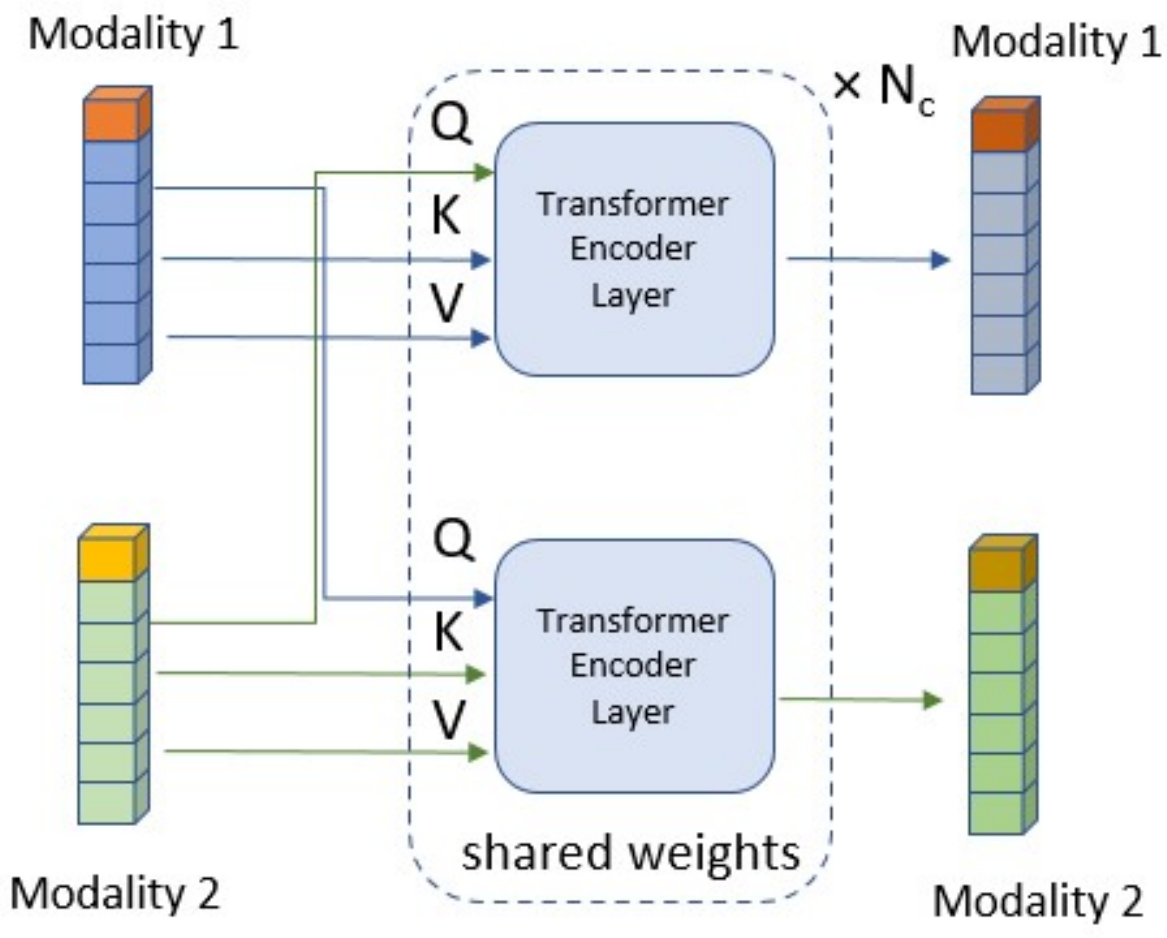}} 
   
\end{tabular}
% \subcaptionbox{Self Attention}

% \subcaptionbox{Cross-Self Attention}{\includegraphics[width=0.4\linewidth]{images/cross-self-attention.pdf}} 

% \vspace{-10pt}
 \caption{Cross-attentional module used in our architecture. We use two cross-attentional layers, i.e., $N_c=2$ in our best model.}
\label{fig:capsules-layer}
\vspace{-5pt}
\end{figure}

\section{Additional Implementation details} \label{extra-impl-details}
Here, we discuss the additional design choices used in our model. The capsules use routing to agree or disagree about the presence of certain entities in the input image. This decision is independent of the question. For instance, if an image has a \textit{bus} and an \textit{elephant} in it, the question cannot affect what is in the image. To this end, the capsule routing is performed only once in our method. However, depending on the question being asked, we may require selecting different entities from the image. We achieve this by doing text-based capsule selection at each layer. The capsule selection layer $\phi$ (equation 3 in the main paper) has shared weights for all encoder layers. We use 8 16GB AMD GPUs for pretraining; finetuning for GQA is performed on a single 16GB GPU. To pretrain with 48 capsules, the batch size of 640 is used.

\section{Additional Architectural details} \label{add-arch-details}
\noindent \textbf{Language Encoder:} 
The language encoder $L_e$ is composed of $L$ transformer encoder layers. Its input is a tokenized sentence $S_l$ of length $l$. 
% Each encoder layer has $h$ heads, $d$ dimension size, and $L$ encoder layers. 
% The set of language tokens $S_l$ is appended with special token $[CLS]$ at the first position and $[SEP]$ at the last index of the sequence. 
The language encoder $L_e$ takes the set of words tokens $\{[CLS], w_1, w_2, ..., w_l, [SEP]\}$ as input, and outputs feature representations $\{h^i_{cls}, h^i_{T_1}, h^i_{T_2}, ..., h^i_{T_l}, h^i_{sep}\}$ at every $i^{th}$ layer, where $h^i_{T_k}$ denotes the text feature for the $k^{th}$ input word token ($k \in {1, 2, ...,l}$) from layer $i$.  Intermediate encoder layer $i$ takes output of previous layer $(i-1)$ as input. $[CLS]$ token is used as the sentence embedding in transformers \cite{devlin2018bert}. Additionally, we use it for capsule selection in visual encoder (section 3.3 in main). 
% We use sentence embedding $[CLS]$ to guide the visual encoder as explained in the next section. 

\noindent \textbf{Visual Encoder:}
The visual encoder $V_e$ has the same architecture as the language encoder with the same dimension size and number of layers. The image embeddings $X'$ are transformed to visual capsules encodings  $X_c$ and input to the visual encoder.  Intermediate layers of visual encoder takes selected visual capsules as residual connection to keep the capsule representation intact while training the system. The final features output $ h^L_{v_j}$ of the visual encoder is used for token-level cross-modality interactions in future steps. Where, $ h^L_{v_j}$ is the feature output for $j^{th}$ visual token ($j \in {1, 2, ..., hw}$) from the last layer $L$. 
% Instead of adding capsules directly to the intermediate layers, we first do capsule selection using the sentence embedding output from the respective layer in the text encoder.

\noindent \textbf{Feature Pooling} 
The feature pooling layer takes text-based features and image-based features as input and outputs a $d$ dimensional feature. This output feature can be used as a pooled output for image-text matching and VQA tasks. The feature pooling layer is a fully connected layer followed by a $tanh$ activation layer. We pretrain our system in two stages before finetuning for VQA task. To be specific, during first stage pretraining , the input is the concatenated features $[h^L_{cls}, h^L_{img}]$ for special tokens from text and image encoders; where $h^L_{cls}$, and $h^L_{img}$ are used as aggregated features over text input and image input respectively. Let $f_P$ be the feature pooling layer, the pooled feature output $h_{{1}_{pooled}}$ will be as follows: 
\begin{equation} \label{eq:7}
h_{{1}_{pooled}} = f_P([h^{L}_{cls}, h^{L}_{img}]) 
\end{equation}

During second stage pretraining, the concatenated features pair after cross attention is indicated as $ [h^{\hat{L}}_{cls}, h^{\hat{L}}_{img}]$ and the pooled feature output is denoted by $h_{{2}_{pooled}}$. The equation is as follows:
\begin{equation} \label{eq:8}
    h_{{2}_{pooled}} = f_P(h^{\hat{L}}_{cls}, h^{\hat{L}}_{img}]) 
\end{equation}

 \subsubsection{Cross-Attention Module}
%  A self attention transformer encoder takes queries (Q), keys (K), and values (V) from the same input sequence. 
%  We experimented with three types of cross attentional blocks: 
%  \noindent \textbf{1. Cross Attention ($X1$):} 
 Given two input feature sequences (output $ h^L_{T_k}$ from text encoder and $ h^L_{v_j}$ from image encoder), cross-attention module is a co-attentional transformer which applies attention from one feature sequence to the other by taking queries from first sequence and keys and values from the second sequence, and vice versa. 
%  \noindent \textbf{2. Self Attention ($X2$):} $X2$ is a self attention transformer encoder layer which takes a concatenated sequence of feature sequences output from the text encoder and the image encoder. 
%  \noindent \textbf{3. Cross-Self Attention ($X3$):} Cross-Self attention ($X3$) is a combination of $X1$ and $X2$ where a cross attention transformer block is followed by a self attention transformer encoder. The output of $X1$ is two feature sequences which are concatenated and then input to the self-attention layer.
 Multiple layers of these cross-attention blocks can be stacked. 
%  Cross-Self attention block, the output is a single sequence from self attention, which is split into feature sequences corresponding to the image and text tokens, and fed to the next layer.
The final text output feature $ h^{\hat{L}}_{cls}$ corresponding to the $[CLS]$ token and final visual output feature $ h^{\hat{L}}_{img}$ corresponding to the $[IMG]$ token are used for pretraining and finetuning the model. Where, $N_c$ is the number of layers in cross-attention module and $\hat{L}=L+N_c$ denotes the depth of the model in terms of number of layers.

\section{Training objectives} \label{train-objectives}

\noindent \textbf{Masked Language Modeling} Masked Language Modeling is a self-supervised language modeling task where a small percentage of words are masked before giving the sentence as input to the language encoder. The task is to predict the masked words using the context from other words in the sentence. This self-supervised approach is very effective to learn strong text representations \cite{devlin2018bert}. The features output $ h^L_{T_k}$ from language encoder $L_e$ is used for training on this task. In the second stage, instead of predicting missing words from solely text features, the masked word is predicted from the visual-guided language features i.e., we take features outputs $ h^{\hat{L}}_{k}$ from the last text-based cross-attention layer. 
% This task is used only in the second stage of pretraining as the first stage of pretraining does not have the cross-attentional blocks. 

\noindent \textbf{Image-Text Matching (ITM)} 
% Given the image-text pair, 
% both inputs are input to their respective transformer encoders. The output embedding against the special tokens ([CLS] for sentence, [IMG] for image) are then concatenated, and given as input to a classifier. 
To predict whether the input pair of image-text features is a matching pair or not, we take the output $h_{{1}_{pooled}}$ (eq. \ref{eq:7}) from feature pooling layer and input to a fully connected layer which outputs logits for each class: 'matching' or 'non-matching'. At the second  pretraining stage, the output features corresponding to $[IMG]$ and $[CLS]$ token after cross-attentional module (each of dimension $d$) are used for prediction. The pretraining head now uses $h_{{2}_{pooled}}$ (eq. \ref{eq:8}) as input for image-text matching task.

\noindent \textbf{Visual Question Answering}
% We use MSCOCO and Visual Genome for pretraining our systems. 
Inspired by \cite{tan2019lxmert}, we also use VQA as one of our pretraining tasks.
% VQA datasets based on MSCOCO or Visual Genome images can be used as a pretraining task i.e. 
We use Visual7W \cite{zhu2016visual7w}, GQA \cite{hudson2019gqa} and VQA \cite{goyal2017making} in our pretraining. Like ITM, we take the pooled features from text and visual encoders and input to a classifier. The classifier is comprised of two fully connected layers. An activation function and layer norm is used between the two layers. The final output is probability scores for each answer.
In the first stage of pretraining, $h_{{1}_{pooled}}$ is used as the pooled feature.
For second stage pretraining, pooled cross-modal feature output $h_{{2}_{pooled}}$ is used for answer prediction. 
A separate softmax cross-entropy loss function is used to optimize each of the above heads. We give equal weights to each loss term during pretraining.

\noindent \textbf{Finetuning parameters for the baselines.}
To finetune LXMERT, we use the same training parameters as our method. ViLT is finetuned with batch size of 256 with lr=$1e-5$. ALBEF is finetuned with batch size of 16. Learning rate is increased to $2e-5$ to speed up training for ALBEF. ALBEF and ViLT use the maximum batch size which could fit in the GPU memory. All models are trained on GQA for upto 10 epochs.

\section{Grounding Performance Evaluation} \label{grounding-eval}
\subsubsection{Choice to use last layer's attention for grounding:}
It is common in the vision-language community to employ the last layer's analysis e.g., DINO\cite{dino} uses the last layer's attention for the object segmentation task without any specialized training objective or architecture.
% We thank R2 for providing the reference which also compare their approach with grad-CAM as well as raw attention maps from the last layer.
% Other examples are GQA [CVPR'19] and MAC-Caps [CVPR'21] to report performance on last (mean) attention maps. 
We follow the protocol of previous works in the field for SOTA comparison to allow for a fair evaluation, namely following MAC~\cite{hudson2018compositional}, MAC-Caps~\cite{khan2021reason} with mean (last) attention scores, ALBEF~\cite{ALBEF} by using the 8th and last layer with Grad-CAM (GC) and attention scores (ATN) and ViLT~\cite{vilt} for the last layer cosine (cos) and attention scores (see Tab. 2 in the main paper) and achieve SOTA grounding performance. However, there is a possibility that some intermediate layer does the better job at grounding such as ALBEF finds that layer 8 in their model is good at grounding. Nevertheless, searching for the best layer is expensive in terms of time and computational cost. Our approach outperforms on grounding even when evaluated for the last layer only. This choice also eliminates the need to search for the best grounding layer within each model and well suited to test the systems for unseen data. \\ 
%--------------------------------------------------------
\begin{table}[t]
\small
\renewcommand{\arraystretch}{.9}
  \centering \setlength{\tabcolsep}{.7\tabcolsep}   
    \begin{tabular}{lccccccc}
       \toprule
                % \multicolumn{1}{c}{\textbf{Method}} & \multicolumn{3}{c}{}  \\
                % \cmidrule(lr){2-4}
      
        & \multicolumn{3}{c}{Overlap} & & \multicolumn{3}{c}{IOU} \\
    \cmidrule{2-4} \cmidrule{6-8} 
   \multicolumn{1}{c}{\textbf{Obj. label}}
     &  P & R & F1 & & P & R & F1 \\
      
        \midrule 
 Answer (A) & 17.64 & 89.81 & 29.49 && 2.05 & 10.45 & 3.42 \\
 Question (Q) & 49.57 & 81.86 & 61.75 && 4.01 & 6.48 & 4.95  \\
    
	\bottomrule
    \end{tabular}
 \vspace{7pt}
    \caption{ GradCAM results for our model. Compare to Tab. 2 in the paper.
    % The top 3 rows show results for models trained with 16 capsules.
    }
    \vspace{-5pt}
    \label{tab:gc_ours}
\end{table} 

  %-------------------------------------------------------------

\noindent \textbf{Ours + Grad-CAM:} 
% Following the \textcolor{purple}{R2}'s suggestion, 
To compare with ALBEF, we  also evaluated our system with Grad-CAM output of the last layer. We observe $\approx2-4\% \uparrow$ increase in overlap F1-score for both Q \& A and a ($\approx 0.5-1.01\% \downarrow$) decrease in IOU F1-score still achieving better grounding results than the baselines (see Tab. \ref{tab:gc_ours} and Tab. 2 (main)).

\subsubsection{Generating heatmaps using ViLT demo:} The demo code visualizes word-to-patch attention for the matching image-caption pair. 
% omitting the sentence-embedding ([CLS] token).
For VQA grounding, we consider question-to-image attention (attention from [CLS] token to visual tokens). The provided code for optimal transport algorithm leads to numerical instability for the question token ([CLS]) generating NaN. 
% The demo code also generates first text token before visualization. 
Hence, we used the cosine scores (computed before optimal transport) as well as raw attention to generate heatmaps. Heatmaps are generated with the same post processing as provided in the demo. To verify this, ($<question\_id>$, $<image\_id>$) pairs for Fig.4 (main) are:
{ \small \{(`00798998', `2356417'), (`02451905', `2386586'), (`00653991', `2324955'), (`00511505', `2410567'), (`01782610', `2409395')
\}}.

\begin{figure*}[t]

\begin{tabular}{ccc}
\subcaptionbox{}{\includegraphics[width=0.3\linewidth]{ 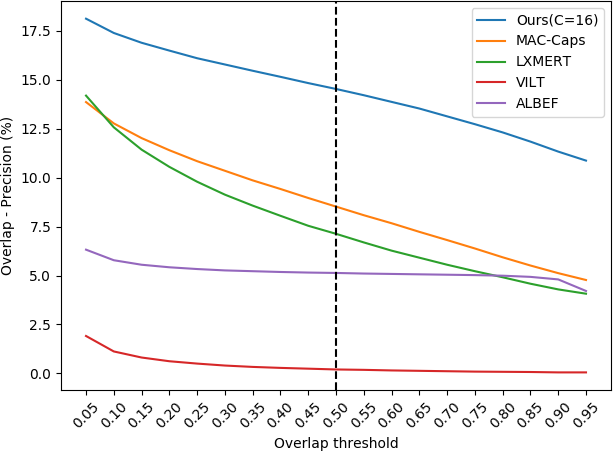}} &
\subcaptionbox{}{\includegraphics[width=0.3\linewidth]{ 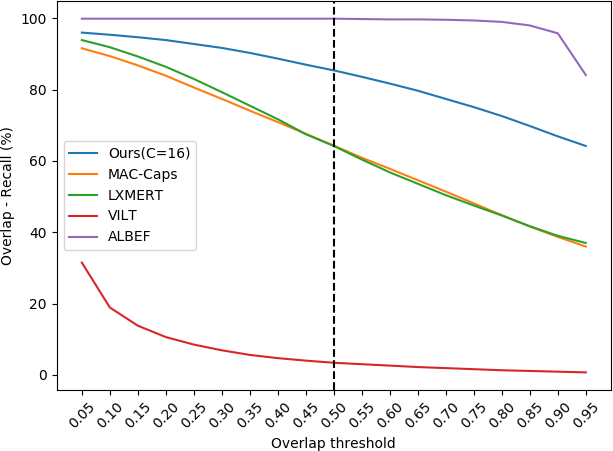}} &
\subcaptionbox{}{\includegraphics[width=0.3\linewidth]{ 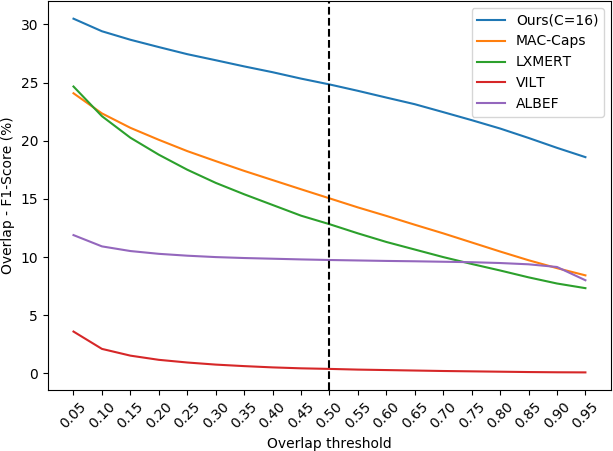}} \\

\subcaptionbox{}{\includegraphics[width=0.3\linewidth]{ 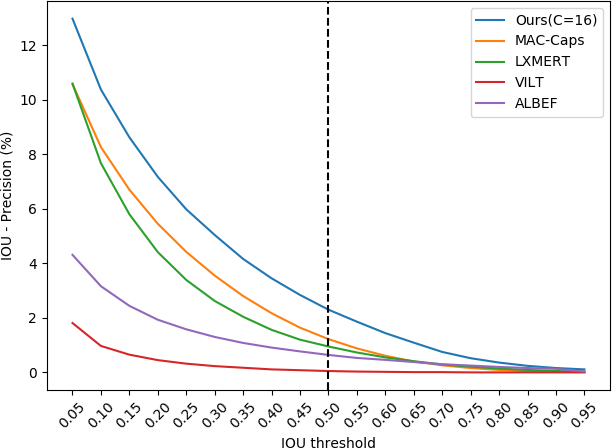}} &
\subcaptionbox{}{\includegraphics[width=0.3\linewidth]{ 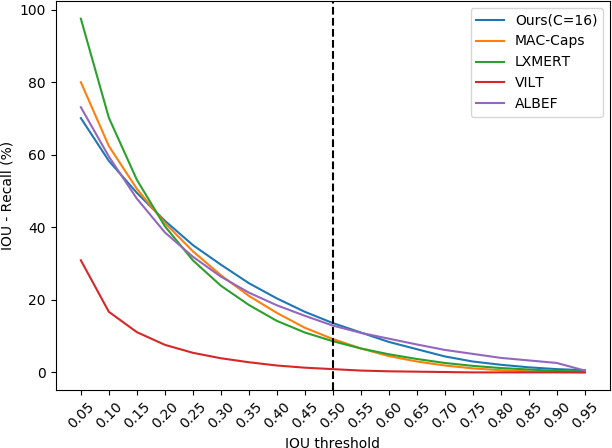}} &
\subcaptionbox{}{\includegraphics[width=0.3\linewidth]{ 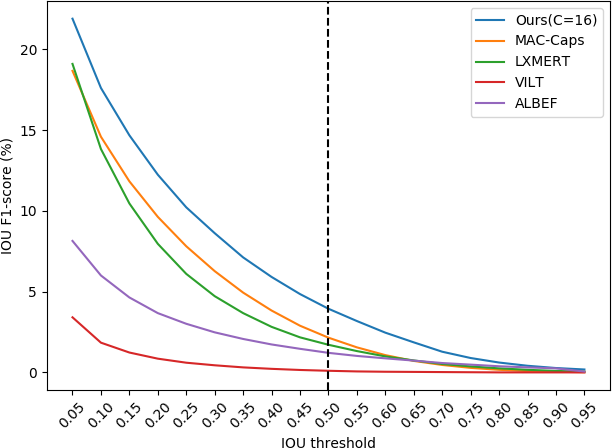}}

\end{tabular}
\vspace{-5pt}
 \caption{\small Comparison with baselines for varying overlap and IOU detection threshold from 0.05 to 0.95. Plots (a), (b), and (c) show the results for varying detection threshold for overlap in terms of precision, recall, and F1-score respectively. Plots (d), (e), and (f) are the results for comparison when varying IOU threshold in terms of precision, recall, and F1-score respectively. Our method is significantly outperforming the baselines for all values of overlap thresholds in terms of precision, and subsequently F1-scores. For IOU, the proposed method is doing well for threshold values as high as 0.8 in terms of precision and F1-score, whereas, IOU-Recall is comparable to ALBEF. }
\label{fig:vary-thresh}
\vspace{-5pt}
\end{figure*}
%--------------------------------------------------------------------------------

\begin{figure*}[t]

\begin{tabular}{ccc}
\subcaptionbox{}{\includegraphics[width=0.3\linewidth]{ 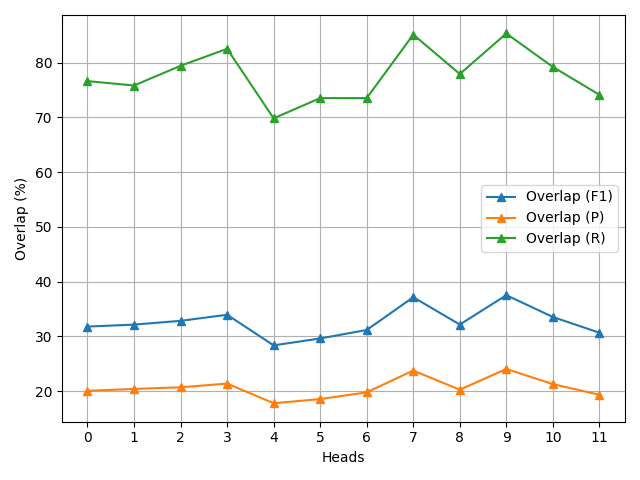}} &
\subcaptionbox{}{\includegraphics[width=0.3\linewidth]{ 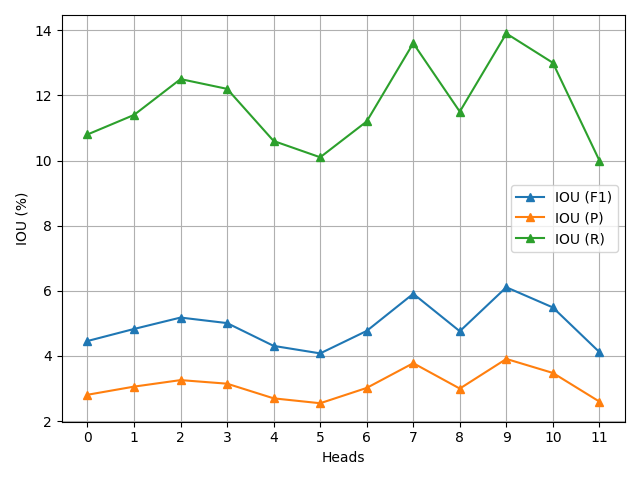}} &
\subcaptionbox{}{\includegraphics[width=0.3\linewidth]{ 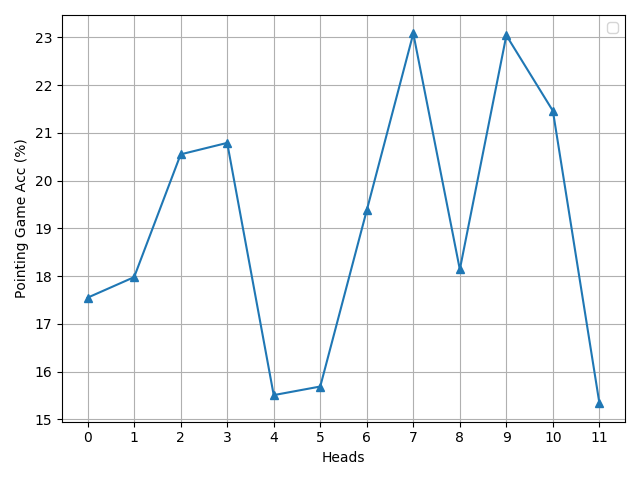}}

\end{tabular}
\vspace{-5pt}
 \caption{\small  Grounding performance from the proposed model (C=16) for each head in the last cross-attention layer. (a) reports overlap accuracies in terms of precision, recall, and F1-score; (b) shows IOU in terms of precision, recall, and F1-score; and (c) shows pointing game accuracy for each head.
 Overall, head 7 and head 10 show best grounding performance among all heads. For pointing game, head 7 achieves the highest accuracy of 23.08\%. Using the proposed way to evaluate the pointing game performance, i.e., clustering over maximum points from all heads, improves pointing game accuracy significantly (34.59\%). }
\label{fig:per-head-grounding}
\vspace{-5pt}
\end{figure*}

%---------------------------------------------------------------------------------
\begin{table*}
\scriptsize
\renewcommand{\arraystretch}{.9}
  \centering \setlength{\tabcolsep}{.2\tabcolsep}   
    \begin{tabular}{lclcccccccc}
       \toprule
                % \multicolumn{1}{c}{\textbf{Method}} & \multicolumn{3}{c}{}  \\
                % \cmidrule(lr){2-4}
      
      &   & & &\multicolumn{3}{c}{Overlap} & & \multicolumn{3}{c}{IOU} \\
    \cmidrule{5-7} \cmidrule{9-11} 
   \multicolumn{1}{c}{\textbf{Q Type}} & Example & Method & Acc.
     &  P & R & F1 & & P & R & F1  \\
      
        \midrule 
    %  (1)  no caps.(X1)     & 62.43 & 11.88 & 65.42 & 20.11 && 1.77 & 9.80 & 3.00 & 34.22 \\
     % (2)  w/o mask       & 95.64 & - & - & - && - & - & - & -\\
        \multirow{2}{*}{Open} & \multirow{2}{*}{\textit{\tiny How is the weather in the image?}} & no-caps & 62.43 & 23.03 & 46.02 & 30.70 && 4.83 & 9.66 & 6.44 \\ 
    %  & no-skip (C=32) & 39.15 & 77.36 & 51.99 && 4.82 & 9.64 & 6.43\\ 
    %   & Ours (C=32)& 36.27 & 72.12 & 48.26 && 4.96 & 9.92 & 6.61\\
   &  & ours& 56.65 & \textbf{43.06} & \textbf{85.57} & \textbf{57.29}  && \textbf{6.62}  & \textbf{13.24} & \textbf{8.83} \\
      \hline \\
    %  (3)  e2e(C=32,X1)    & 57.02 & 10.36 & 64.28 & 17.84 && 1.33 & 8.30 & 2.29 & -\\
    %  (4)  2-stage(C=32,X1) & 55.41    & 10.09 & 71.95 & 17.70 && 1.41 & 10.09 & 2.47 & -\\
     %(5)  2-stage, freeze wts(X1)  & - & - & - & - && - & - & -& -\\
     \multirow{2}{*}{Binary} & \multirow{2}{*}{\textit{\tiny Is it cloudy today?}} & no-caps & 62.43 & 33.19 & 66.10 & 44.20  && 7.13 & 14.25 & 9.50 \\
    % & no-skip (C=32) & 39.44 & 77.99 & 52.39 && 7.89 & 15.79 & 10.52\\
    %   & Ours (C=32)& 31.67 & 63.22 & 42.20  && 7.53 & 15.07 & 10.04\\
  &  & ours& 56.65 & \textbf{42.54} & \textbf{84.67} & \textbf{56.63} && \textbf{12.12} & \textbf{24.23} & \textbf{16.15}\\
        \midrule 
         \midrule 
       \multirow{2}{*}{Query} & \multirow{2}{*}{\textit{\tiny What kind of fruit is on the table?}} & no-caps & 62.43 & 33.19 & 66.10 & 44.20 && 4.83 & 9.66 & 6.44 \\
    %  & no-skip (C=32) & 39.15 & 51.99 & 77.36 && 4.82 & 9.64 & 6.43\\
    %   & Ours (C=32)& 36.27 & 72.12 & 48.26 && 4.96 & 9.92 & 6.61 \\
  &   & ours& 56.65 & \textbf{43.06} & \textbf{85.57} & \textbf{57.29} && \textbf{6.62} & \textbf{13.24} & \textbf{8.83}\\
     \hline \\
     \multirow{2}{*}{Compare} & \multirow{2}{*}{\textit{\tiny Who is taller, the boy or the girl?}}& no-caps & 62.43 & 9.96 & 19.91 & 13.27 && 1.11 & 2.21 & 1.47 \\
    %  & no-skip (C=32) & 40.71 & 80.35 & 54.04 && 2.43 & 4.87 & 3.24\\
    %   & Ours (C=32)& 25.22 & 50.00 & 33.53 && 1.99 & 3.98 & 2.65\\
  &   & ours& 56.65 & \textbf{37.83} & \textbf{75.00} & \textbf{50.29} && \textbf{2.43} & \textbf{4.87} & \textbf{3.24}\\
     \hline \\
       \multirow{2}{*}{Choose} & \multirow{2}{*}{\textit{\tiny Is it sunny or cloudy?}} & no-caps & 62.43 & 24.62 & 49.19 & 32.82 && 7.86 & 15.72 & 10.48 \\
    %  & no-skip (C=32) & 39.29 & 77.70 & 52.19 && 8.56 & 17.12 & 11.41\\
    %   & Ours (C=32)& 32.45 & 64.84 & 43.26 && 8.21 & 16.42 & 10.94\\
  &   & ours& 56.65 & \textbf{43.11} & \textbf{85.85} & \textbf{57.40} && \textbf{13.29} & \textbf{26.59} & \textbf{17.73}\\
     \midrule 
     \midrule
     \multirow{2}{*}{Category} & \multirow{2}{*}{\textit{\tiny What kind of fruit is it, an apple or a banana?}} & no-caps & 62.43 & 30.14 & 60.10 & 40.14 && 8.66 & 17.32 & 11.54 \\
    %  & no-skip (C=32) & 37.46 & 74.28 & 49.81 && 7.50 & 15.00 & 10.00\\
    %   & Ours (C=32)& 34.65 & 69.04 & 46.14 && 8.43 & 16.85 & 11.23\\
  &   & ours& 56.65 & \textbf{42.90} & \textbf{85.49} & \textbf{57.13} && \textbf{11.73} & \textbf{23.46} & \textbf{15.64}\\
     \hline \\
    %  (14) bert init (C=32, X1)  & - & - & - & - && - & - & - & -\\
     \multirow{2}{*}{Relation} & \multirow{2}{*}{\textit{\tiny Is there an apple on the black table?}}  & no-caps & 62.43 & 33.45 & 66.60 & 44.53 && 4.16 & 8.31 & 5.54 \\
    %  & no-skip (C=32) & 39.52 & 78.03 & 52.47 && 4.42 & 8.85 & 5.90 \\
    %   & Ours (C=32)& 36.44 & 72.44 & 48.48 && 4.37 & 8.75 & 5.83\\
   &  & ours& 56.65 & \textbf{43.10} & \textbf{85.60} & \textbf{57.33} && \textbf{5.84} & \textbf{11.68} & \textbf{7.79}\\
     \hline \\
      \multirow{2}{*}{Attribute} &  \multirow{2}{*}{\textit{\tiny what color is the apple?}} & no-caps & 62.43 & 9.96 & 19.91 & 13.27 && 1.11 & 2.21 & 1.47  \\
    %  & no-skip (C=32) & 40.71 & 80.35 & 54.04 && 2.43 & 4.87 & 3.24\\
    %   & Ours (C=32)& 25.22 & 50.00 & 33.53 && 1.99 & 3.98 & 2.65 \\
  &   & ours& 56.65 & \textbf{37.83} & \textbf{75.00} & \textbf{50.29} && \textbf{2.43} & \textbf{4.87} & \textbf{3.24}\\
    %  \midrule
    %  \midrule
    %  \midrule
    %   \multirow{2}{*}{Overall} & no-caps & - & - & - && - & - & -  \\
    %  & Ours (C=16) & 14.55 & 85.54 & 24.87 && 2.32 & 13.70 & 3.96\\
    \hline
	\bottomrule
    \end{tabular}
%  \vspace{-5pt}
    \caption{\small Comparison of our backbone model with no capsules (no-caps) and the proposed model with 16 capsules (Ours (C=16)). Results are shown w.r.t. each question type. Adding capsules to the backbone model significantly improves the grounding performance for all question types.  
    % The top 3 rows show results for models trained with 16 capsules.
    }
    % \vspace{-15pt}
    \label{tab:ablation_qtype}

\end{table*} 
%----------------------------------------------------------------------

\section{Performance analysis w.r.t. varying detection threshold} \label{vary-threshold}
We use detection threshold=0.5 for all our results in the submitted paper. For overlap, a detection is considered to be a true positive when the overlap between the ground truth box and the predicted region is greater than 0.5. Similarly, a detected region with an IOU of greater than 0.5 over a ground truth bounding box is considered a true positive for IOU. In figure \ref{fig:vary-thresh}, we study the impact of having a very low detection threshold vs. employing high thresholds by varying the threshold from 0.05 upto 0.95. We observe that the proposed method is robust to detection threshold for the overlap metric in terms of precision and F1-score even for the very high threshold of 0.95. For IOU, we also perform well for precision and F1-score. For IOU in terms of recall, our method and ALBEF show comparable results.

\section{Grounding accuracy w.r.t. each head} \label{vary-head}
Grounding accuracy for individual heads in the last cross-attentional layer are shown in figure \ref{fig:per-head-grounding}. The results are reported in terms of precision, recall, and F1-score for overlap and IOU. For pointing game, the maximum point over the attention map produced from each head is used to evaluate the per-head pointing game accuracy. Using clustering over the points obtained from each head outperforms the best performing head by $\uparrow 11.51\%$ (best head: 23.08\% vs. clustering: 34.59\%).

\section{Results w.r.t. question type} \label{qtype}

Table \ref{tab:ablation_qtype} shows grounding results of our best model for different question types. GQA has questions classified with respect to structural type and  semantic type. We compare our model with our backbone model which uses no capsules. Our system outperforms over all question types for both overlap and IOU particularly for question type ``compare'', ``choose'', and ``attribute''. Examples for each question type are provided in table \ref{tab:ablation_qtype}.Refer to GQA \cite{hudson2019gqa} for more details about the question types present in this dataset. 

%-------------------------------------------
\begin{figure*}[t]

\begin{tabular}{ccc}
\subcaptionbox{capsule 1: surfing, outdoor sports}{\includegraphics[width=0.3\linewidth]{ 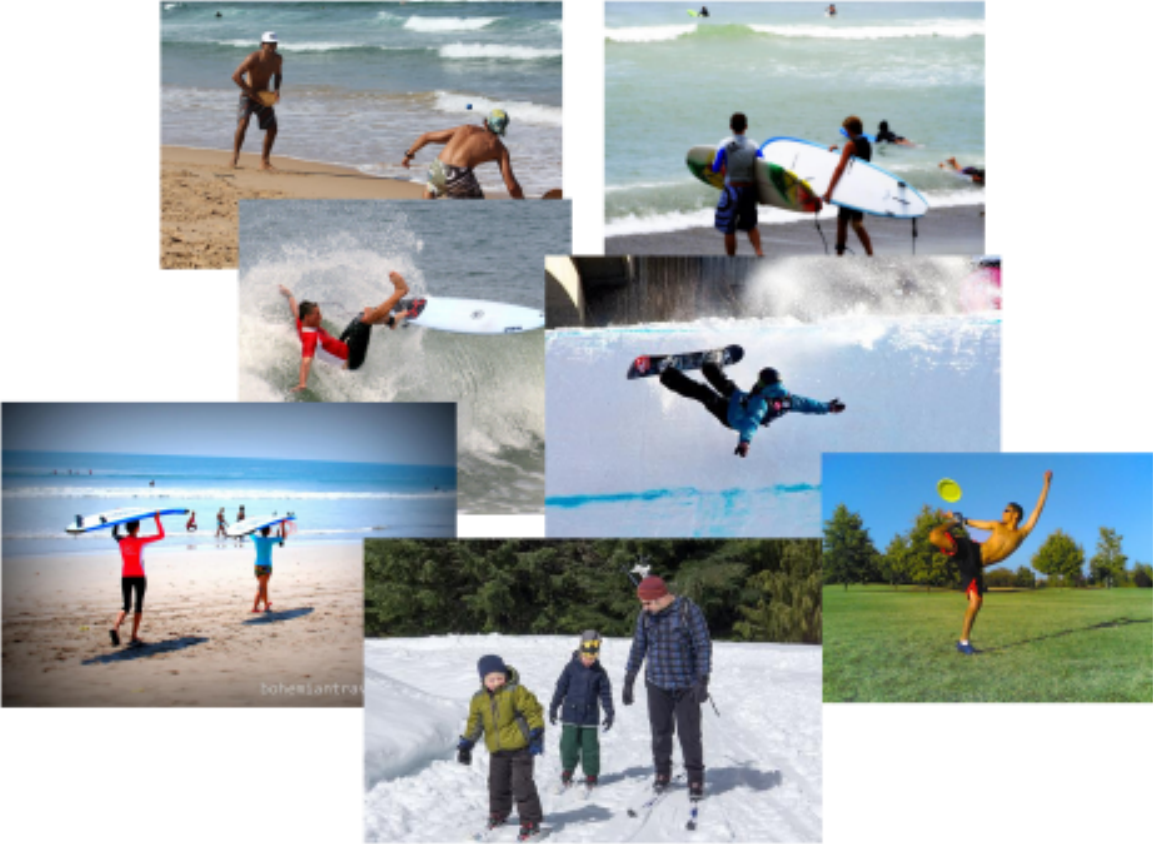}} &
\subcaptionbox{capsule2: food}{\includegraphics[width=0.3\linewidth]{ 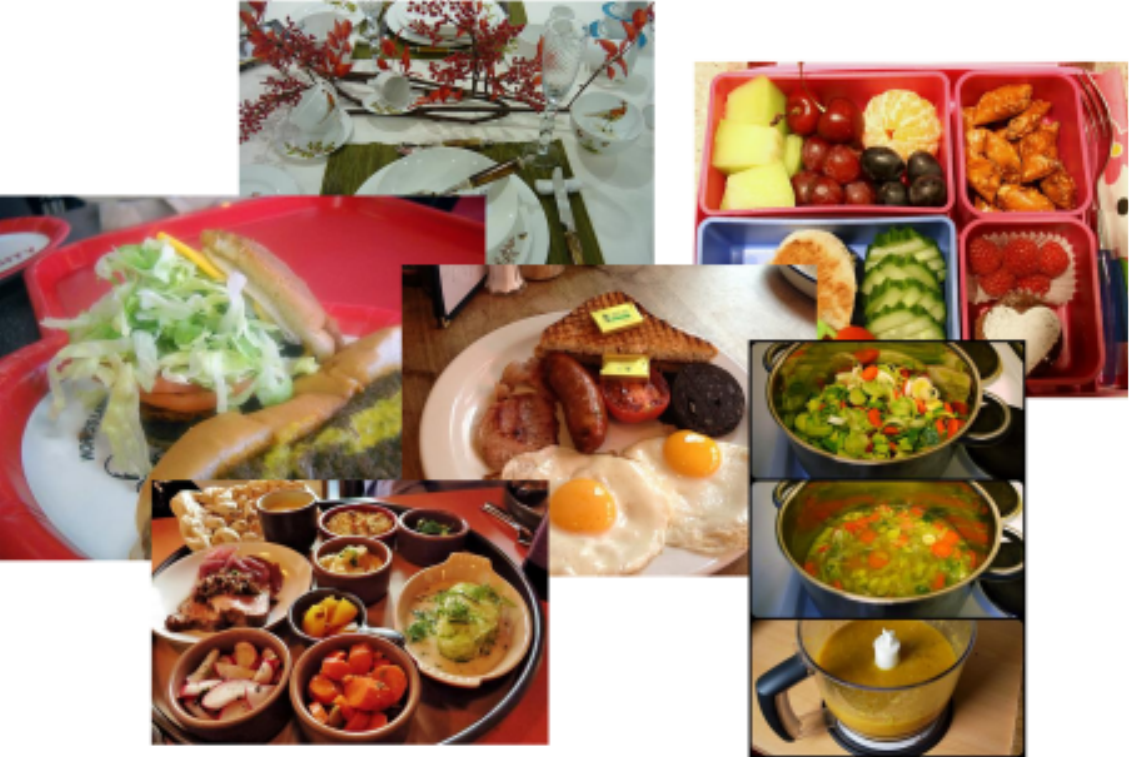}} &
\subcaptionbox{capsule 4: toys, zebra, horse}{\includegraphics[width=0.3\linewidth]{ 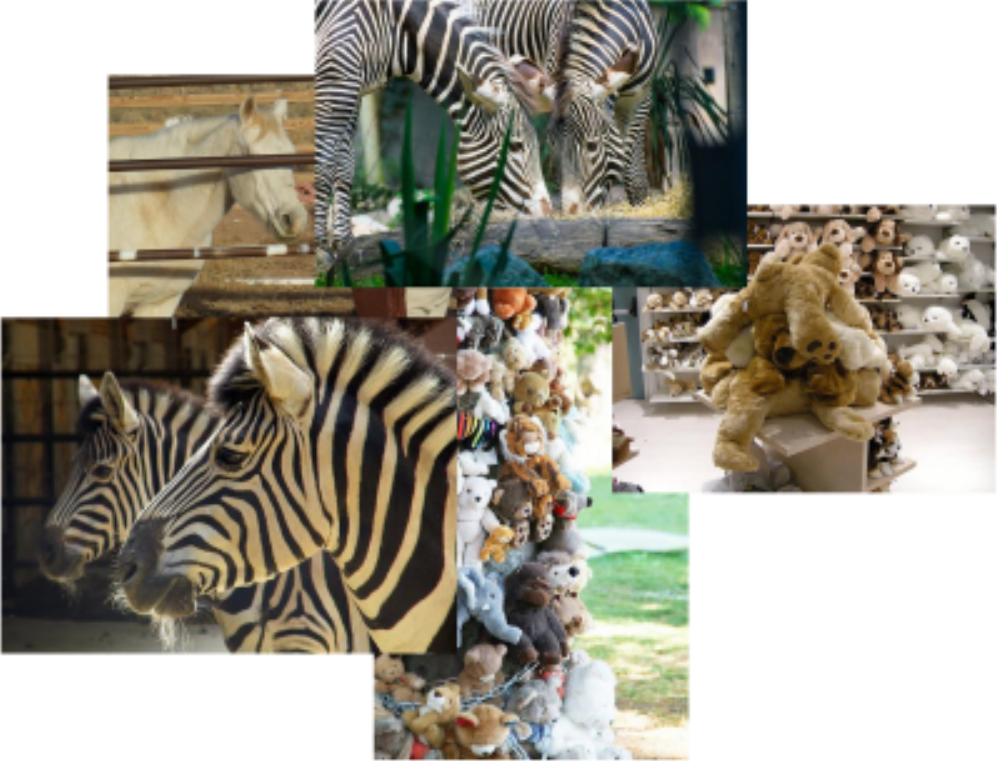}} \\

\subcaptionbox{capsule 5: elephants, wild animals}{\includegraphics[width=0.3\linewidth]{ 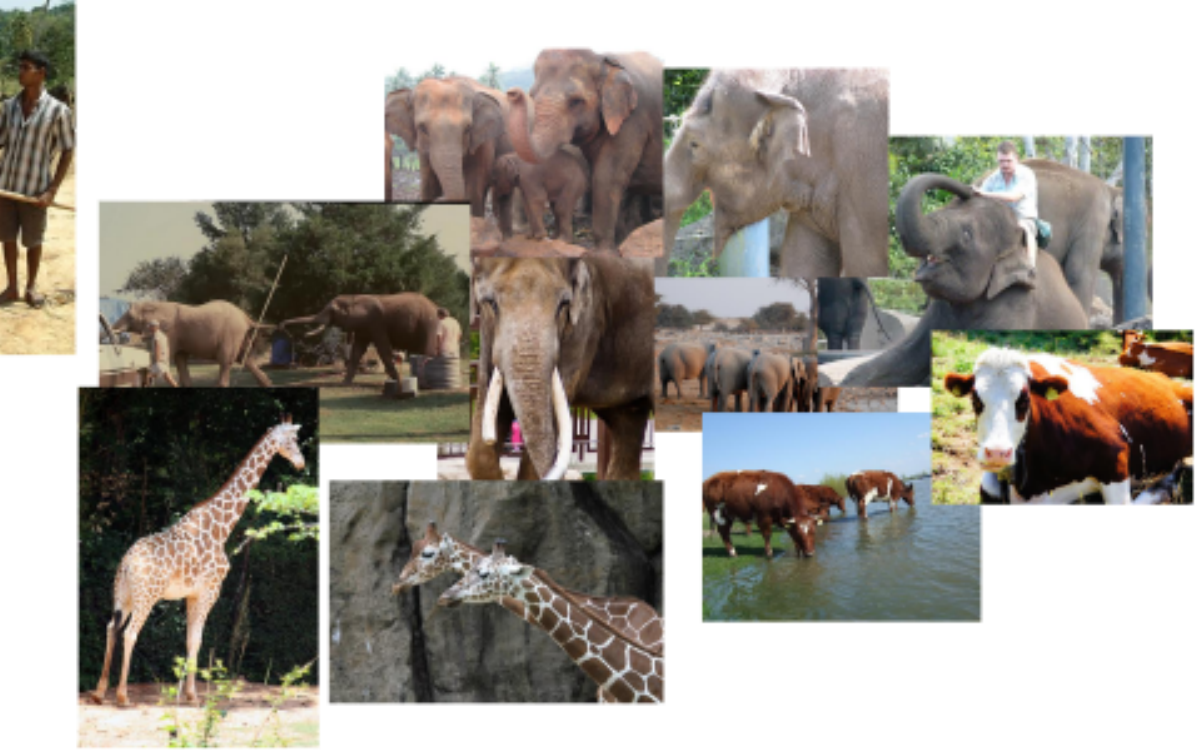}} &
\subcaptionbox{capsule 6: sports, tennis}{\includegraphics[width=0.3\linewidth]{ 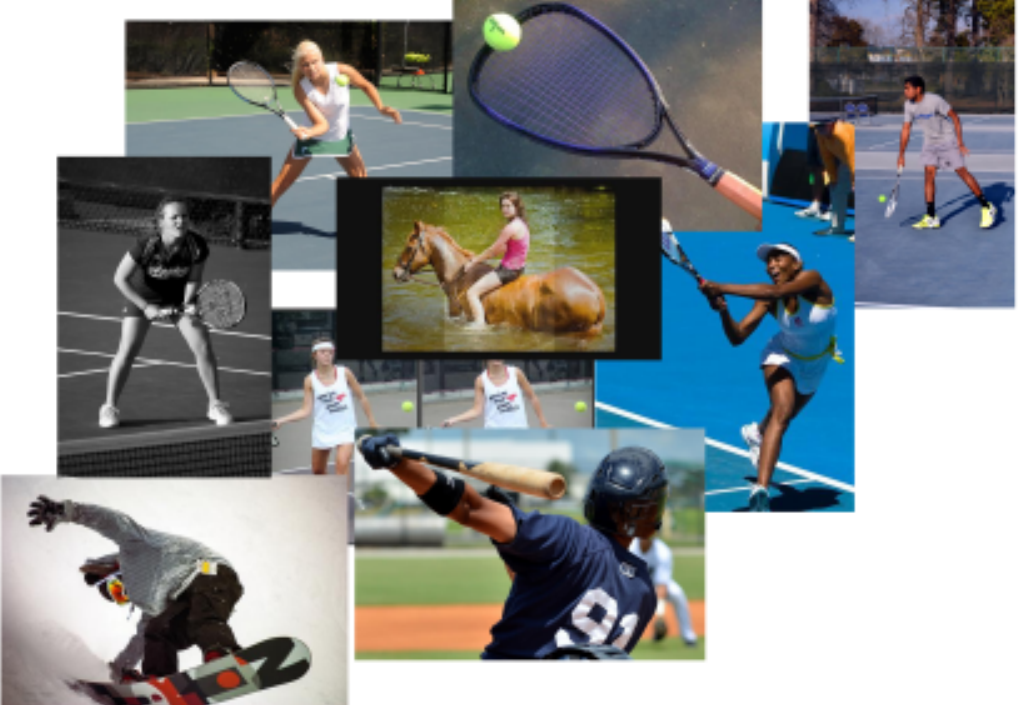}} &
\subcaptionbox{capsule 9: pizza}{\includegraphics[width=0.3\linewidth]{ 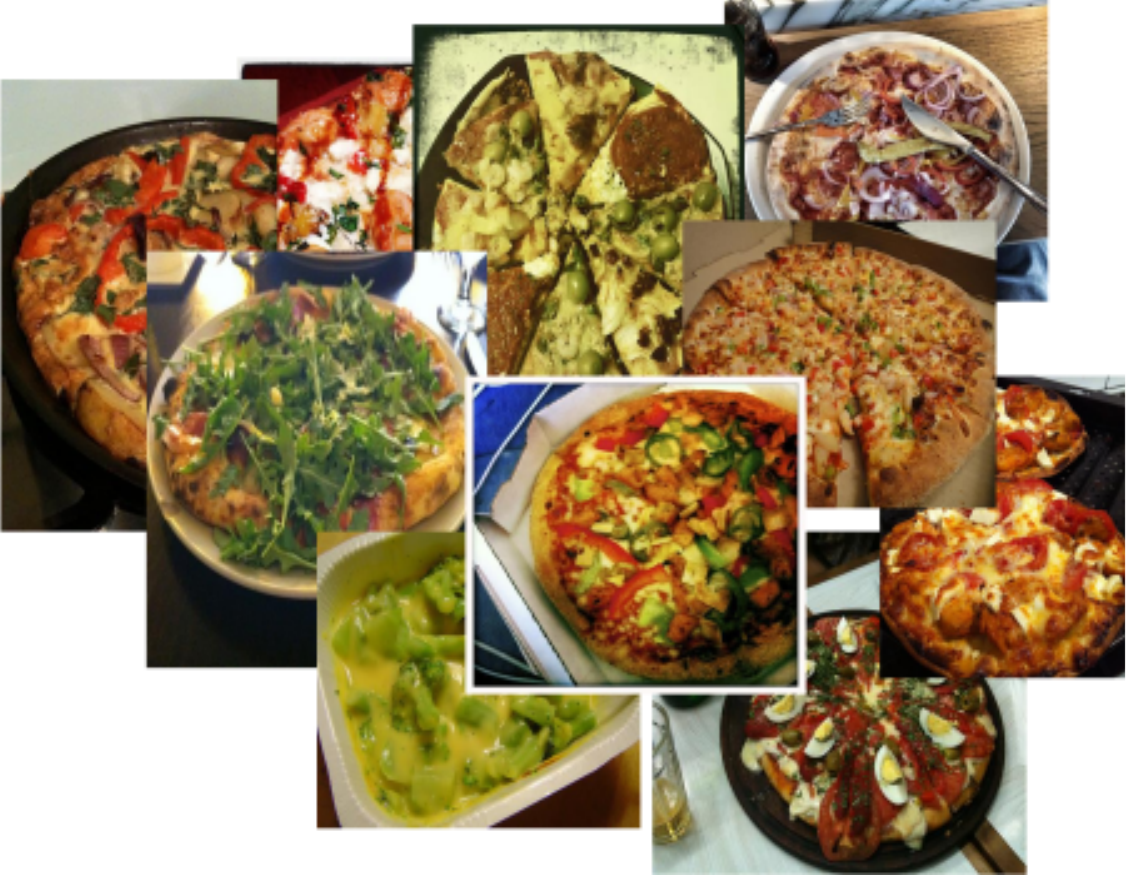}} \\

\subcaptionbox{capsule 11: humans}{\includegraphics[width=0.3\linewidth]{ 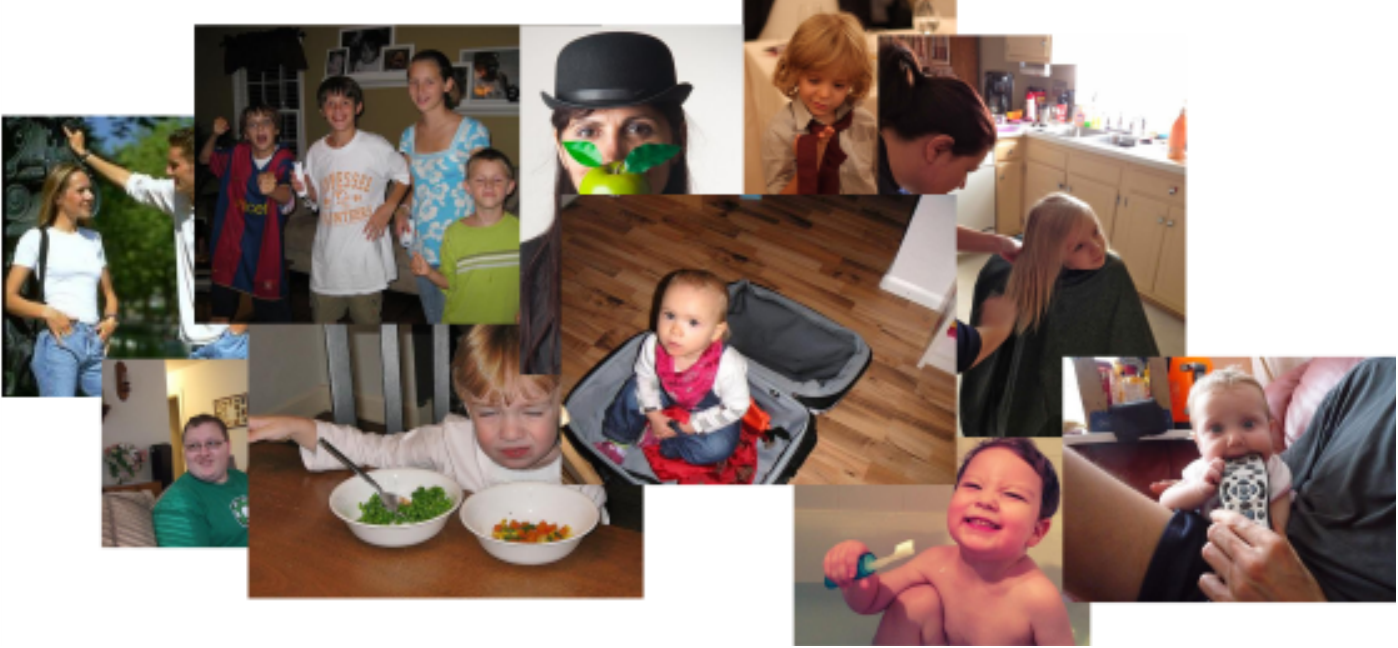}} &
\subcaptionbox{capsule 13: buildings}{\includegraphics[width=0.3\linewidth]{ 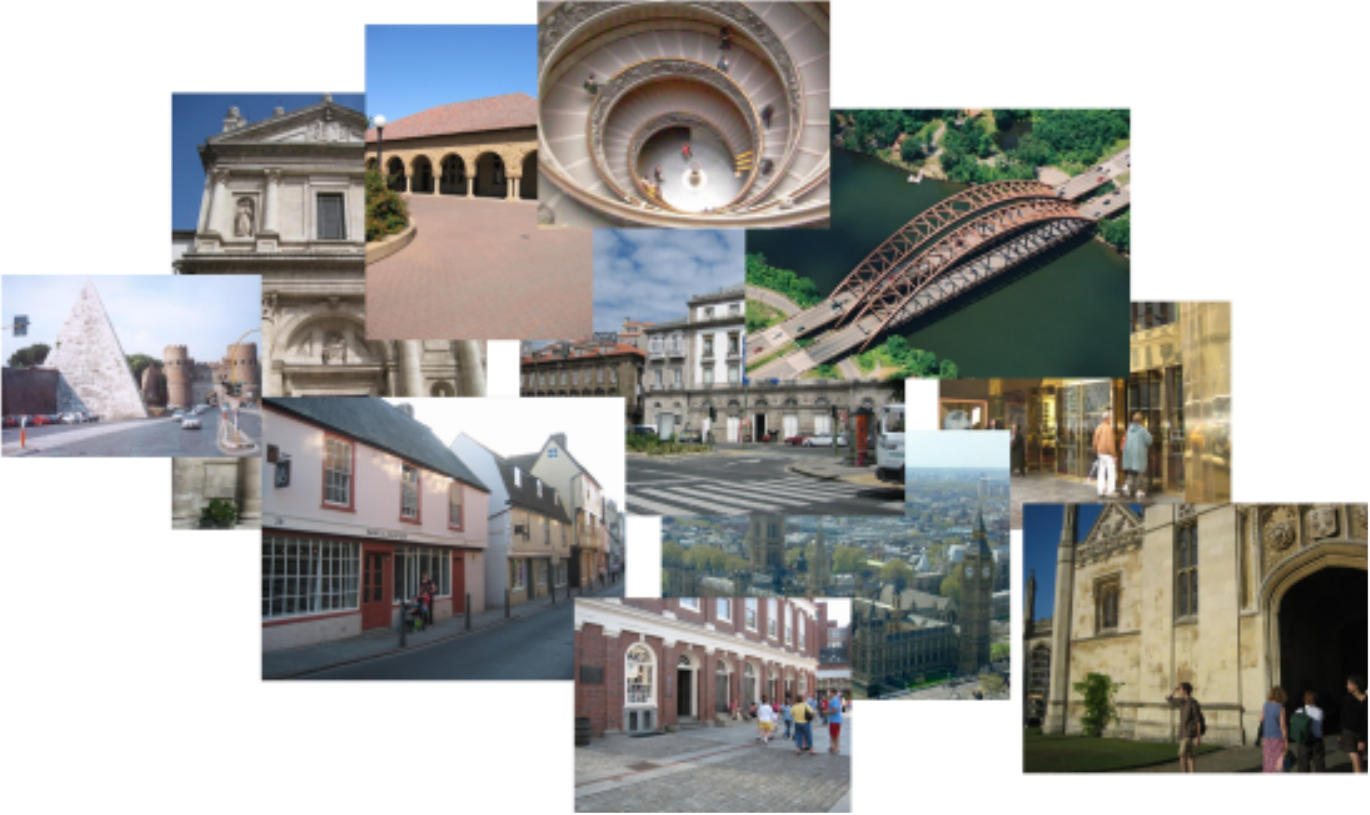}} &
\subcaptionbox{capsule 15: bathroom, trains}{\includegraphics[width=0.3\linewidth]{ 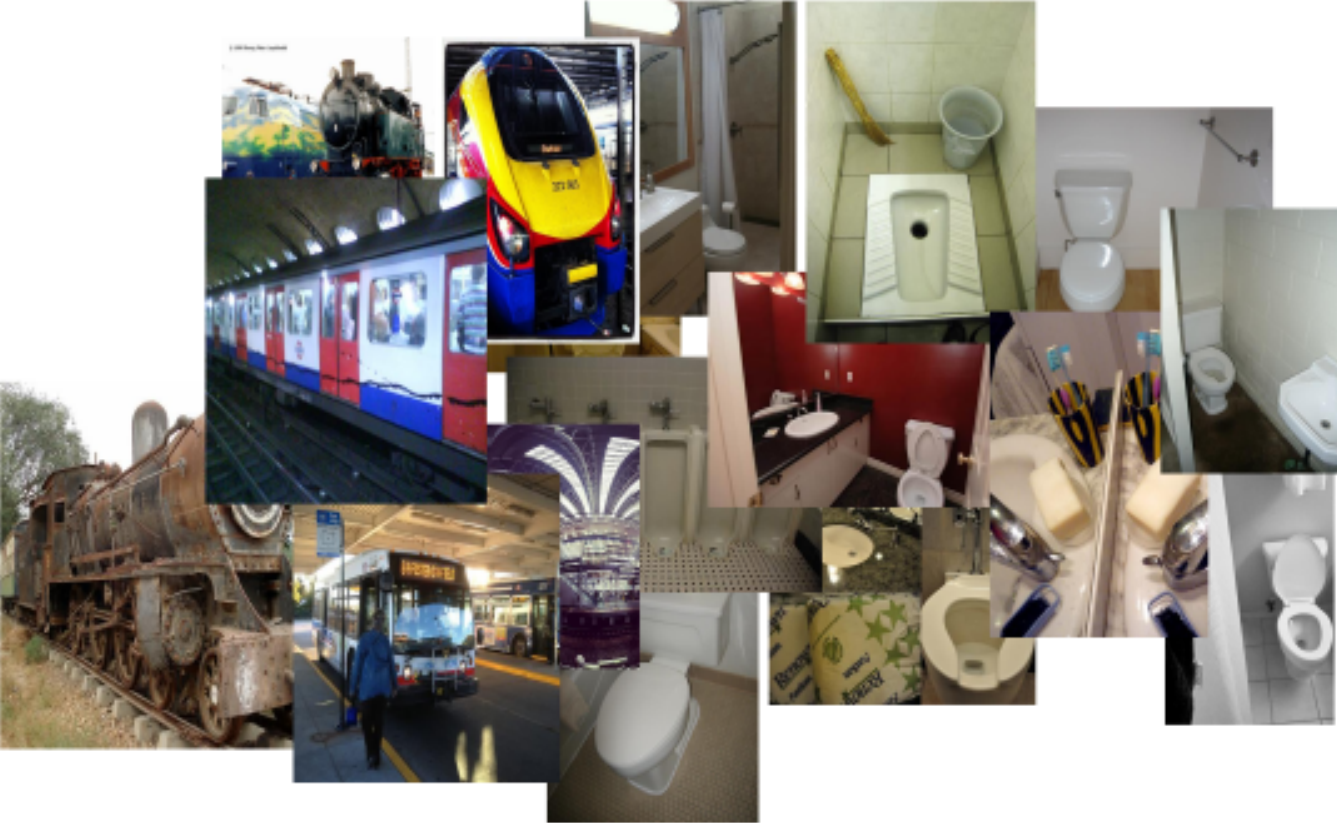}} \\

\end{tabular}
\vspace{-5pt}
 \caption{\small  Images represented by individual capsules. Here, we show the group of images where a certain capsule has the highest activation, e.g., capsule 9 has the highest activation when there is \textit{pizza} in the image. }
\label{fig:capsules}
\vspace{-5pt}
\end{figure*}
%----------------------------------------------------------------------

\section{Entities represented by capsules} \label{capsule-vis}
Since, we do not use class labels to train the capsules, and use VQA supervision instead for training the whole system, it is hard to guess which entity is being represented by each capsule.
To examine what individual capsules are learning, we take the average over capsule activations for each spatial location resulting in a vector of dimension C (C=number of capsules). Each feature in that C-dimensional vector shows the average activation (presence probability) of an individual capsule for that image. The highest activated capsule is used to sort the images into C groups. Figure \ref{fig:capsules} shows the images where a given capsule had the highest activation. In the figure, we can see different capsules are focused on different types of images, e.g., capsule 1 is mostly focused on \textit{outdoor sports} like \textit{surfing}. Since these capsule representations are learned in a weakly-supervised manner, they show  overlapping behavior over certain image classes. Some of them exhibit an interesting behavior. For instance, while capsule 2 is focused on \textit{food items}, capsule 9 is fond of \textit{pizza}; capsule 5 has learned what an \textit{elephant} looks like, but also good at identifying \textit{giraffes} and \textit{cows} in the wild; capsule 13 is focused on \textit{buildings}. We used our best model with C=16 capsules for these visualizations. 
%----------------------------------------------------------------
\begin{table}[t]
\small
\renewcommand{\arraystretch}{.9}
  \centering \setlength{\tabcolsep}{.9\tabcolsep}   
    \begin{tabular}{lcc}
       \toprule

 Method  & depth (\#transformer layers) & \#Params (M) \\

     \midrule  
    % MAC \cite{hudson2018compositional} &  & \\
                        
    % MAC-Caps \cite{khan2021reason} &  &  \\
     LXMERT \cite{tan2019lxmert} & 12 &  239.8\\
     ALBEF \cite{ALBEF} & 12 & 209.5\\%123.7M ($BERT_{base}$) + 85.8 (ViT) \\
     ViLT \cite{vilt} & 12 &  87.4  \\
     \midrule
     Ours & 7 & 141.0\\

    %   Ours - DINO  &  & &\\
   
	\bottomrule
    \end{tabular}
%  \vspace{-5pt}
    \caption{\footnotesize Number of parameters in all transformer-based methods.}
    \label{tab:num_parameters}
% \vspace{-15pt}
\end{table} 

%----------------------------------------------------------------------

\section{No. of training parameters} \label{train-param}
We compare the proposed model with other transformer-based methods in table \ref{tab:num_parameters}. Our proposed system is shallower than the baseline methods using 5 layers in each modality-specific encoders followed by 2 cross-attentional layers. We denote the length of a vertical stack of transformer layers as the model's \textit{depth}. We follow \cite{vilt, tan2019lxmert} and consider one single modality layer as $1/2$ of a multimodal layer. Hence, the proposed model has the depth=7 compared to the baselines with depth=12. In comparison to other transformer-methods, the proposed system uses less parameters ($\approx$ 141M) than LXMERT \cite{tan2019lxmert} (239.8M) and ALBEF \cite{ALBEF} (209.5M). ViLT has the least number of parameters (87.4M). However, it is a single stream model compared to all other two-stream methods considered for this work including the proposed architecture. We excluded the text embedding layer when computing the number of training parameters since it is shared among all vision-language transformers which are used in this study \cite{vilt}.

%--------------------------------------------------------
\begin{table}[t]
\small
\renewcommand{\arraystretch}{.9}
  \centering \setlength{\tabcolsep}{.9\tabcolsep}   
    \begin{tabular}{lccc}
      \toprule

  \textit{Answer Plausibility}   & ALBEF   & ViLT  & Ours \\

     \midrule  
%   Validity (\%)   & 95.14  & 95.17 & 95.14 & \textbf{95.29} & 95.25 & \textcolor{blue}{\textbf{95.27}}\\
 \textit{for all}    &                  
   92.12 &  92.28 & \textbf{92.30}\\
\hline
\hline \\
\textit{for mispredicted} &
    %   Ours - DINO  &  & &\\
     85.14 & 86.35 & \textbf{87.15} \\
   
	\bottomrule
    \end{tabular}
%  \vspace{-5pt}
    \caption{\footnotesize Plausibility comparison on GQA-val set with ALBEF and ViLT for all question-answer pairs and the mispredicted question-answer pairs. We perform on par (even slightly better) than the baselines in terms of the predicted answer's plausibility. This verifies the system is predicting reasonable answers in the real-world context.}
    \label{tab:plausibility-full}
% \vspace{-10pt}
\end{table} 
%----------------------------------------------------------------------

\section{VQA accuracy of ours vs. baselines:} \label{vqa-acc}
% \auk{
% \begin{enumerate}
%     \item Our method is producing equally valid and plausible answers despite of low accuracy (show comparison with baselines on Validity and Plausiblity)
%     \item Show results when image is noise/random to show the SOTA has language bias
% \end{enumerate}
% }
% \textbf{Different accuracy because of pretraining:}
Our proposed system while achieving better VQA accuracy than previous grounding SOTA on GQA dataset (MAC-Caps ~\cite{khan2021reason}]) and LXMERT (a transformer-based model with object-detection), performs lower than ViLT and ALBEF. We attribute this to two reasons: 1) Less training data -- ViLT and ALBEF are using SBU and GCC additionally with strong data augmentations, so we assume that using additional data and comparable resources for training  would improve our accuracy as well. % We propose to add ViLT (and ALBEF) results trained with same regime in the camera-ready version. 
% Regarding the lower accuracy with respect to VQA in more detail, we found that even while we're getting lower overall accuracy, we perform on par with other methods with respect to validity and plausibility.
% Validity measures that system should give a valid answer e.g., a color for a ``what color" question type. 
 2)  Considering failure cases in more detail, we find that the lower accuracy is mainly driven by semantically correct, but literally wrong answers such as girl vs women or herd vs cow (see examples in Fig. \ref{fig:qualitative2}, \ref{fig:qualitative4}, and \ref{fig:qualitative6}). The answer prediction despite being reasonable is incorrect in terms of language mismatch with the ground truth. It could be possible that capsules help to prevent dataset biases, as they regularize and constrain the training and therefore suppress "shortcuts" based on dataset noise. To validate this further, we compute the plausibility metric for all questions as well as incorrect predictions. \textbf{Plausiblity} measures that an answer is reasonable in the real-world context e.g., it is unlikely to see a `blue' apple in real-world.
%  Our results are on par.
 We perform on par with ViLT and ALBEF for all predicted answers. When compared on the mispredicted questions for all three methods, our system predicts 2\% more plausible answers than ALBEF and 0.8\% better than ViLT (see table \ref{tab:plausibility-full}). 
%  \textbf{All Q}: ALBEF=92.12\%, ViLT=92.28\%, Ours=\textbf{92.30\%}; \textbf{incorrect predicted Q}: ALBEF=85.14\%, ViLT=86.35\%, Ours=\textbf{87.15\%}. 
This study maintains the observation about the predicted answer while being reasonable in the real-world is considered incorrect in terms of exact match with the ground truth consequently leading to decreased VQA accuracy.

\section{Additional details about evaluation on VQA-HAT dataset} \label{vqa-hat}
VQA-HAT dataset provides human attention maps for VQA task. This dataset is based on VQA v1.0 dataset and provides 1473 QA pairs with 488 images in validation set. To evaluate on this dataset, we train our system on VQA v1.0 and evaluate on VQA-HAT validation set. 
The answer vocabulary of VQA train set has a long tail distribution. We follow previous works \cite{antol2015vqa,vqahat} and use 1000 most frequent answers. We first combine training (248,349 QA pairs) and validation data (121,512 QA pairs) to get a total of 368487 QA pairs. We then filter out the questions with out-of-vocabulary answers (answer vocab size is kept 1K)
resulting in 318827 QA pairs. 
We separate out 10K QA pairs from the training set (after above mentioned question filtering) and use it as a validation set to pick our best model. We therefore use 308K QA pairs from VQA v1.0 train and val set for finetuning our pretrained backbone with 16 capsules. The learning parameters used for this training are lr=4e-5, batch size=64, with bert optimizer and trained for 20 epochs. The best model on validation set is used for evaluation.

% %-----------------------------------------------------------
% %  \vspace{-25pt}
% \begin{table}[t]
% \scriptsize
% \renewcommand{\arraystretch}{.9}
%   \centering \setlength{\tabcolsep}{.2\tabcolsep}   
%     \begin{tabular}{lccccccccc}
%       \toprule
%                 % \multicolumn{1}{c}{\textbf{Method}} & \multicolumn{3}{c}{}  \\
%                 % \cmidrule(lr){2-4}

%       &   & \multicolumn{3}{c}{Overlap} & & \multicolumn{3}{c}{IOU} & Pointing Game\\
%     \cmidrule{3-5} \cmidrule{7-9} 
%   \multicolumn{1}{c}{\textbf{Method}} & Acc.
%      &  P & R & F1 & & P & R & F1 &\\
      
%         \midrule 
%  (1) ours, w/o cross  & 59.68 & - & - & - && - & - & - & -\\
%  (2) ours-full & 56.65 & 14.53 & 85.47 & 24.84 && 2.30 & 13.61 & 3.94 & 34.59\\

%     \hline
% % 	\bottomrule
%     \end{tabular}
%  \vspace{-5pt}
%     \caption{\small Effect of adding cross attention module on top of text and image encoders. We notice that without cross attention module (row (1) ours, w/o cross), the task performance is poor but grounding ability is retained. Adding cross-attention module (ours-full) gives us the best of both worlds with improved task accuracy as well as good grounding performance. Both models are trained with 16 capsules.
%     % The top 3 rows show results for models trained with 16 capsules.
%     }
%     \vspace{-15pt}
%     \label{tab:ablations_cross_attn_module}

% \end{table} 
%-------------------------------------------------------------------------

\section{Qualitative Results} \label{qual}
\begin{figure*}
\begin{center}
\includegraphics[width=\linewidth]{ 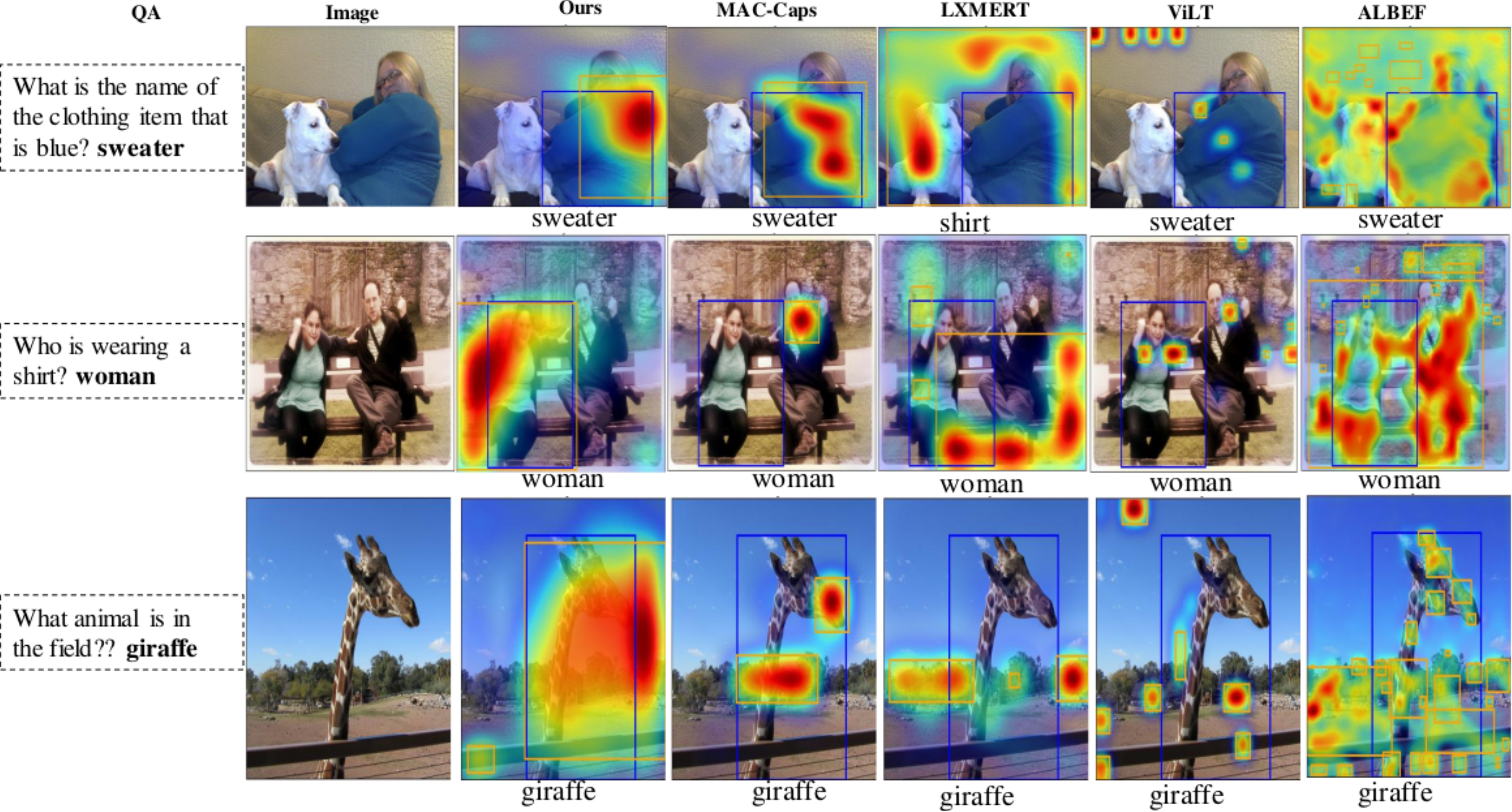}
\includegraphics[width=\linewidth]{ 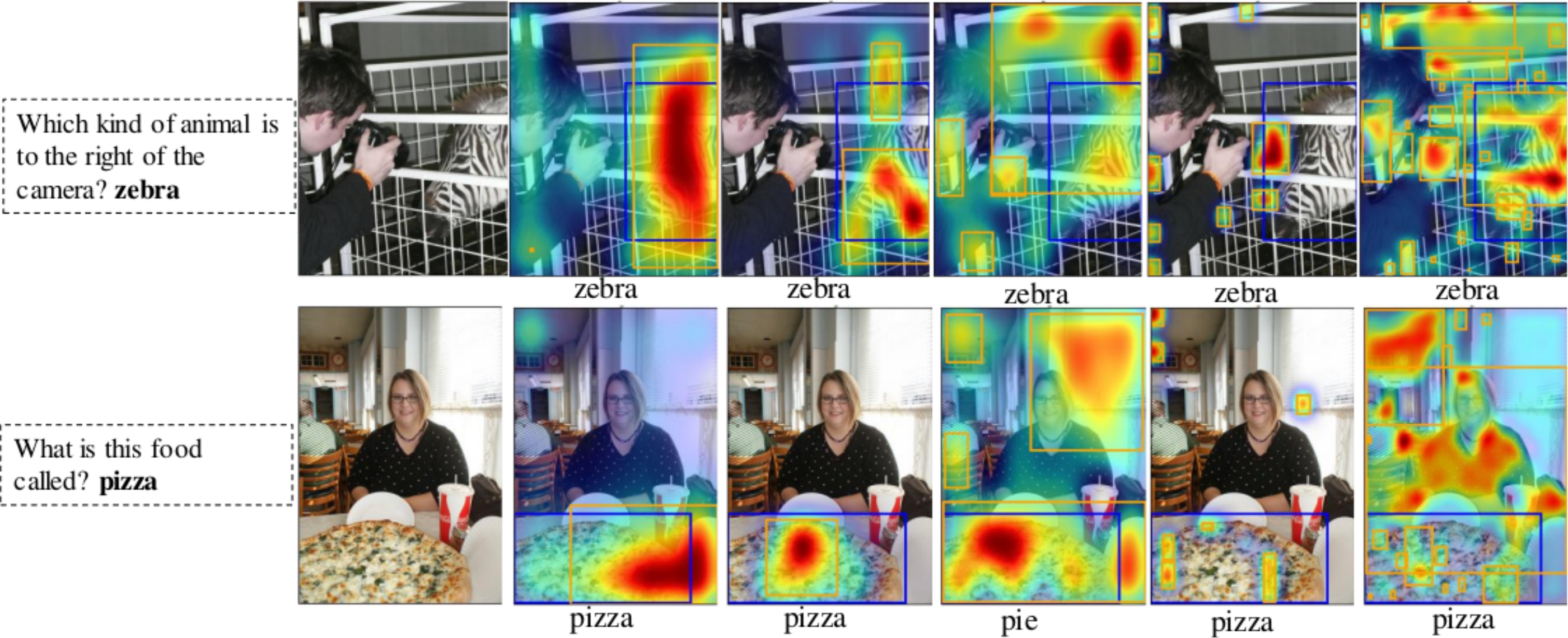}
\includegraphics[width=\linewidth]{ 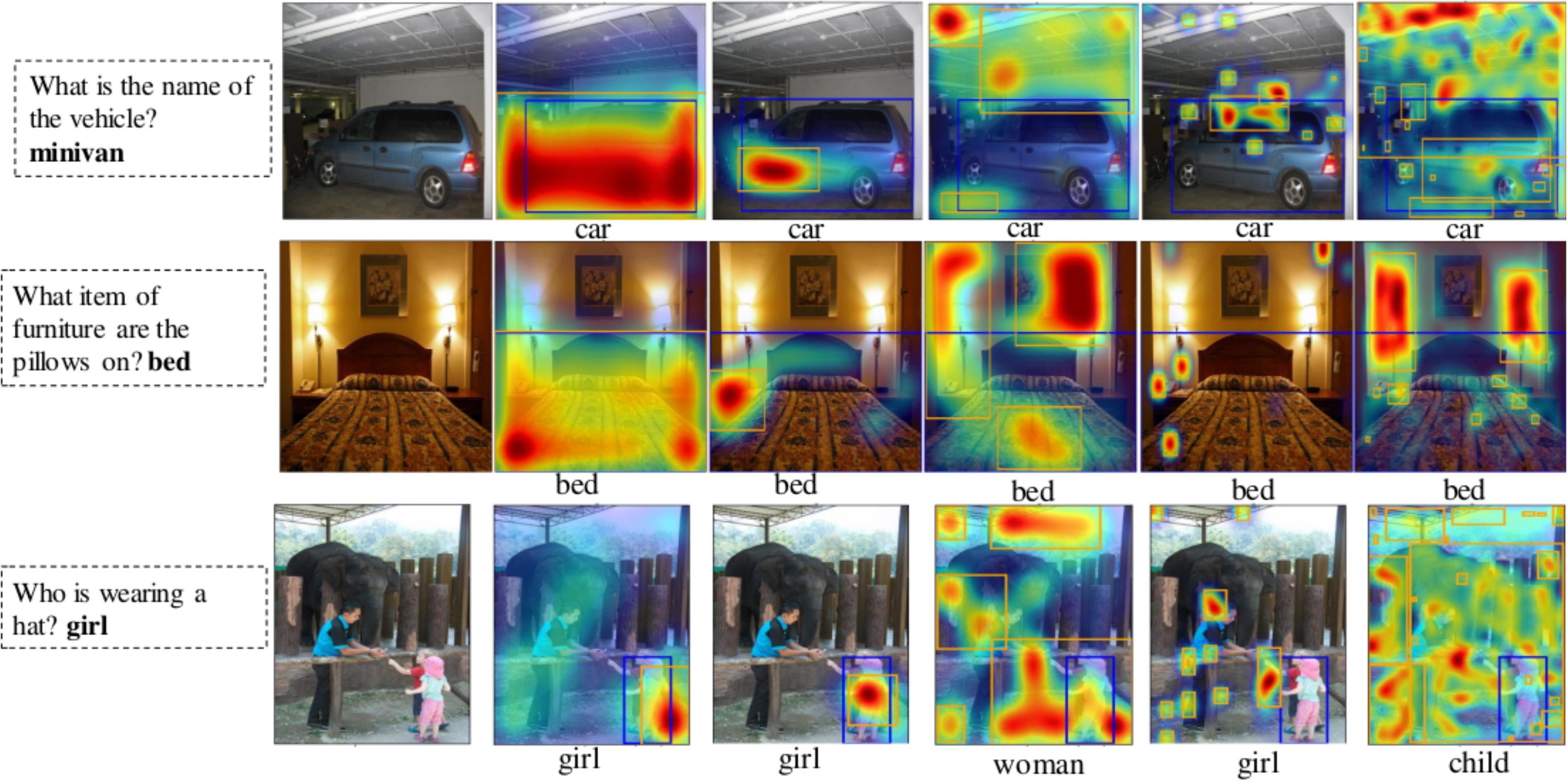}
\end{center}
\vspace{-20pt}
   \caption{\small More qualitative examples where the model predicted the answer correctly with attention (with detected orange boxes) on the correct image regions (blue boxes). Best viewed in color.
   }
   \vspace{-15pt}
\label{fig:qualitative2}
\end{figure*}

\begin{figure*}[t]
\begin{center}

\includegraphics[width=\linewidth]{ 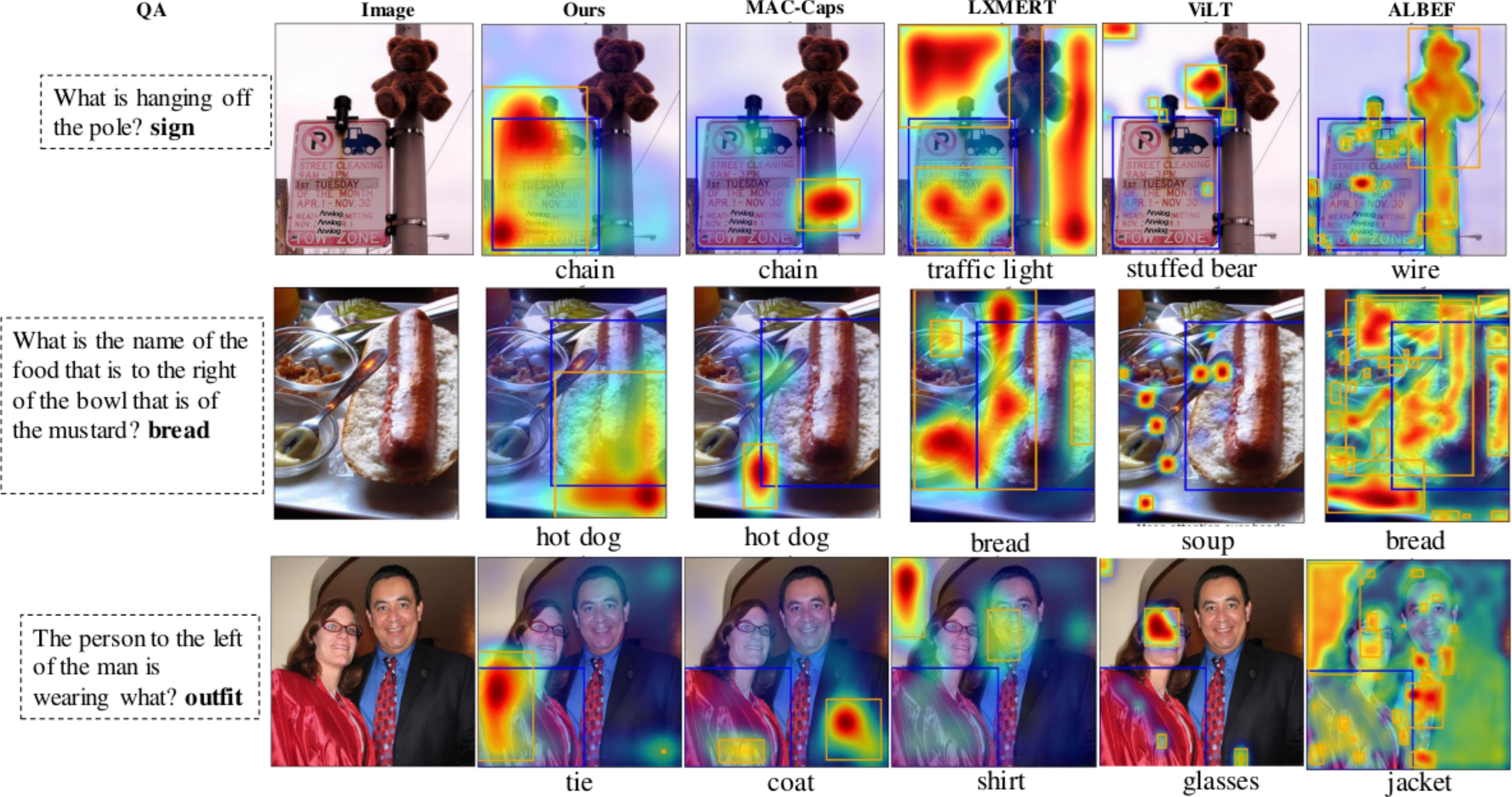}
\end{center}
\vspace{-20pt}
   \caption{\small Qualitative comparison for examples where our model predicted the wrong answer. The attention is over the correct image region. 
   }
   \vspace{-15pt}
\label{fig:qualitative3}
\end{figure*}

\begin{figure*}
\begin{center}

\includegraphics[width=\linewidth]{ 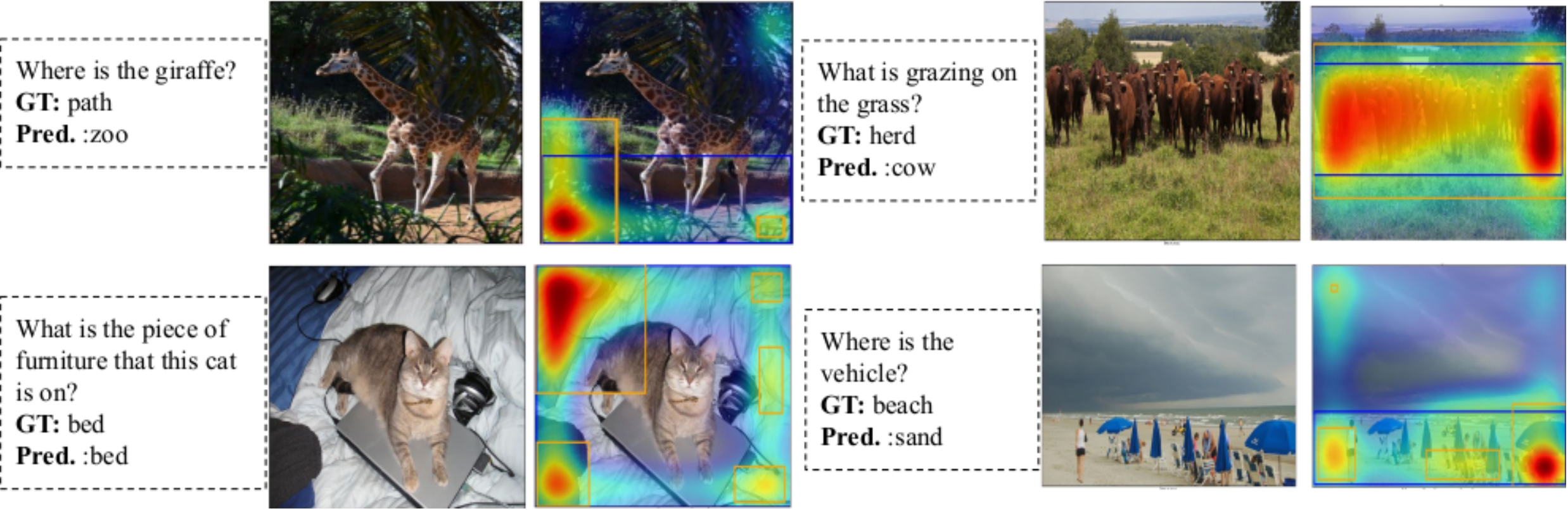}

\end{center}
% \vspace{-20pt}
   \caption{\small Some failure cases for our system. Left two examples show the failure of grounding, right two examples are failure cases in terms of answer prediction. However, both the grounding output and the answers are plausible.  
   }
%   \vspace{-15pt}
\label{fig:qualitative6}
\end{figure*}

\begin{figure*}[t]
\begin{center}

\includegraphics[width=\linewidth]{ 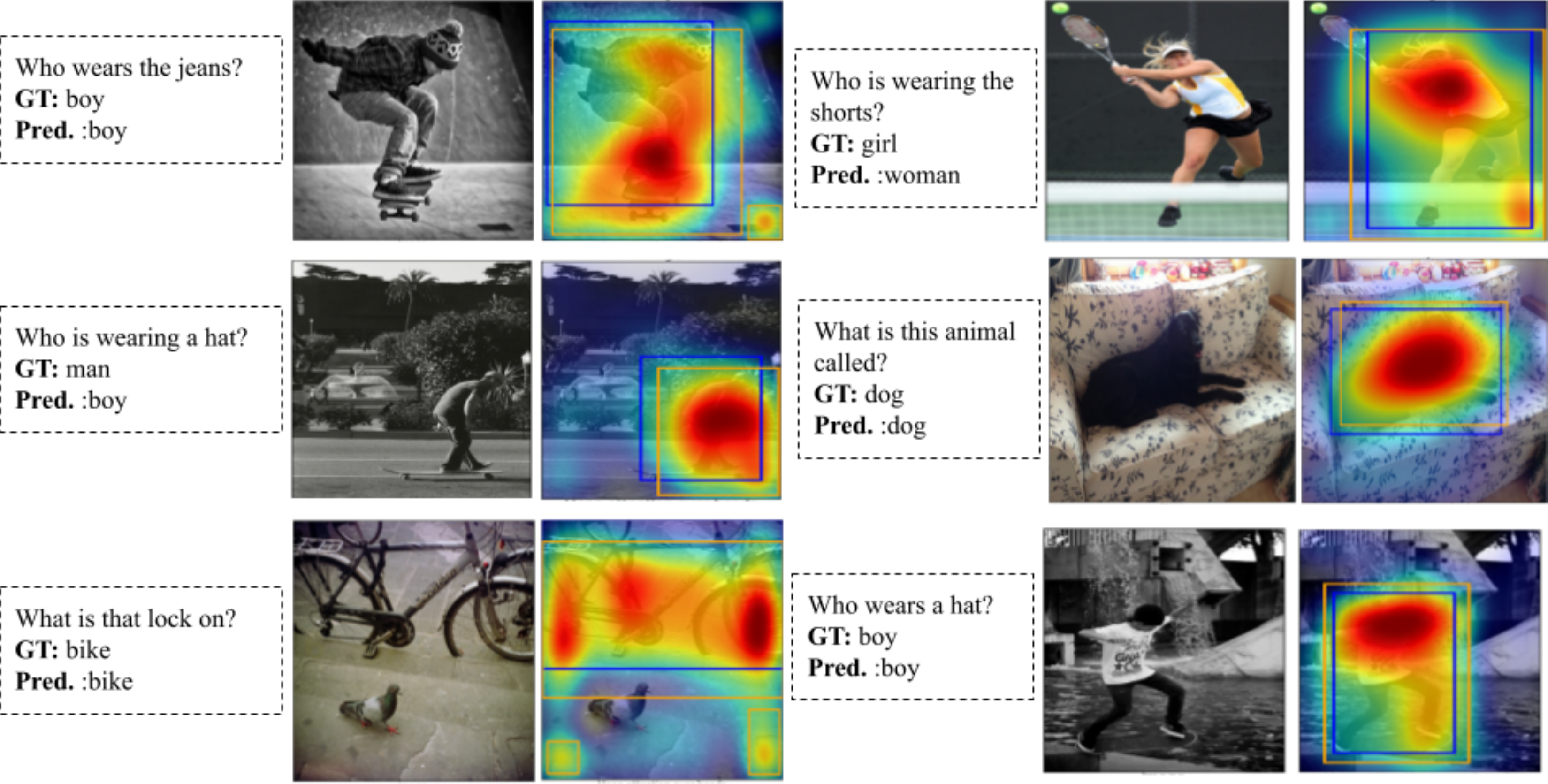}
\includegraphics[width=\linewidth]{ 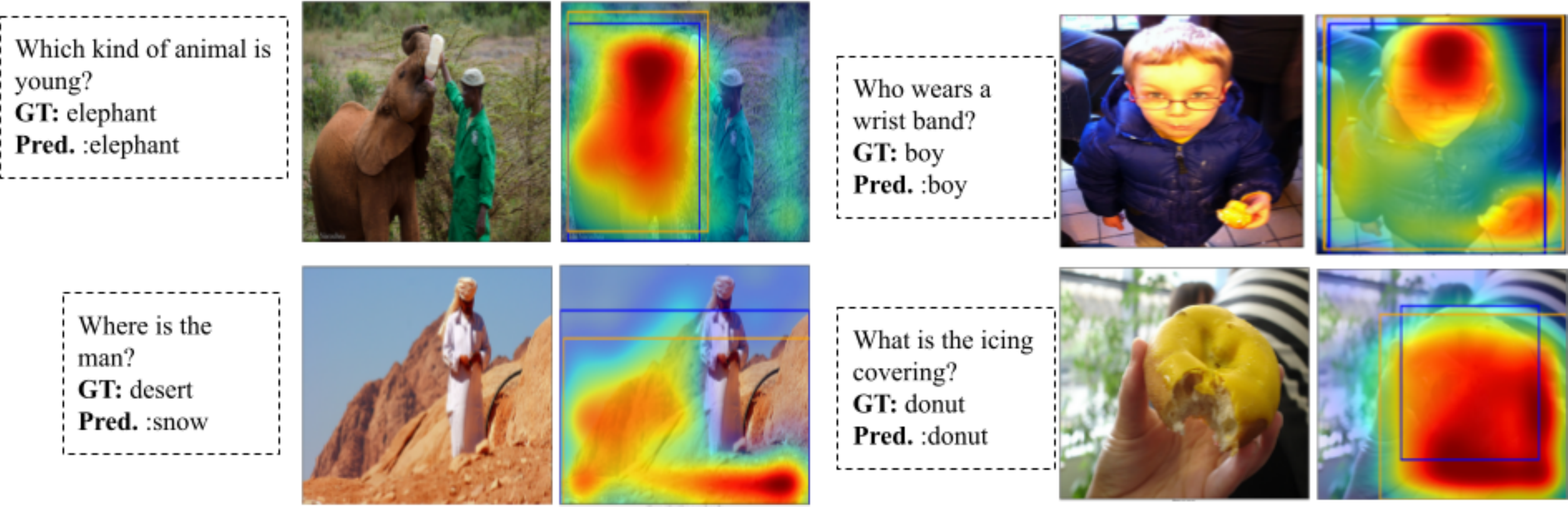}
\end{center}
% \vspace{-20pt}
   \caption{\small More qualitative examples from our system.  
   }
%   \vspace{-15pt}
\label{fig:qualitative4}
\end{figure*}

In figure \ref{fig:qualitative2} and \ref{fig:qualitative3}, we show more examples for qualitative comparison with baselines. Our system consistently produces correct grounding attention when compared to the baselines. In figure \ref{fig:qualitative6}, we show some failure cases for our system in terms of grounding output as well as answer prediction. For grounding failure (in terms of IOU with the groundtruth box), we observe that the system's attention is cogent. For instance, in the top left example, for the question \textit{where is the giraffe?}, the system is looking at the surroundings of giraffe and predicting the answer \textit{zoo}. In the right two examples, the system is grounding correctly even generating reasonable answers. However, these answers are considered incorrect in terms of language mismatch with the groundtruth answer (herd vs. cow and beach vs. sand). Finally, in figure \ref{fig:qualitative4}, we present more qualitative examples from our system.

% \begin{figure*}
% \begin{center}

% \includegraphics{ images/qual-7.pdf}

% \end{center}
% % \vspace{-20pt}
%   \caption{\small More qualitative examples from our system.  
%   }
% %   \vspace{-15pt}
% \label{fig:qualitative5}
% \end{figure*}

\end{document}